\RequirePackage{forest}

\documentclass[sn-basic]{sn-jnl}

\normalbaroutside

\usepackage{grffile}
\usepackage{subfig}

\usepackage{amsmath}
\usepackage{amssymb}
\usepackage{amsthm}
\usepackage{mathtools}
\usepackage{forest}
\usepackage{upgreek}

\usepackage{lscape}
\usepackage[capitalise,noabbrev]{cleveref}
\usepackage{dsfont}
\usepackage{bm}
\usepackage{bbm}
\usepackage{array}
\usepackage{multirow}
\usepackage{multicol}
\usepackage[most]{tcolorbox}
\usepackage{colortbl}
\usepackage{etoolbox}
\usepackage{url}
\usepackage{latexsym}

\usepackage{amsfonts}       
\usepackage{nicefrac}       
\usepackage{microtype}      

\usepackage{xpatch}
\xapptocmd\appendices{%
  \crefalias{section}{appendix}%
}{}{\PatchFailed}

\DeclareMathOperator{\E}{\mathbb{E}}

\DeclareMathOperator*{\argmax}{argmax}

\DeclareMathOperator*{\argmin}{argmin}
\DeclareMathOperator{\KL}{\mathbb{KL}}

\DeclareMathOperator{\Dir}{\mathrm{Dir}}

\DeclareMathOperator{\Gem}{\mathrm{GEM}}

\DeclareMathOperator{\DP}{\mathrm{DP}}

\DeclareMathOperator{\D}{\mathrm{Discrete}}
\DeclareMathOperator{\discrete}{\mathrm{Discrete}}

\DeclareMathOperator{\Beta}{\mathrm{Beta}}

\DeclareMathOperator{\N}{\mathrm{Normal}}
\DeclareMathOperator{\gig}{\mathrm{GIG}}
\DeclareMathOperator{\ig}{\mathrm{IG}}

\newcommand{\m}[1]{\mathbf{#1}}
\DeclareMathOperator{\ncrp}{\mathrm{nCRP}}
\DeclareMathOperator{\crp}{\mathrm{CRP}}

\newcommand{\data}{\m{X}}

\newcommand{\model}{\mathcal{M}}
\renewcommand{\x}{\mathbf{x}}

\newcommand{\tree}{\mathcal{H}}
\newcommand{\edgeset}{E_{\tree}}
\newcommand{\pathset}{\mathbb{Z}_{\tree}}
\newcommand{\nodeset}{Z_{\tree}}

\newcommand{\I}{\m{I}}

\DeclarePairedDelimiterX{\infdivx}[2]{[}{]}{%
  #1\;\delimsize\|\;#2%
}

\newcommand{\parent}{\mathrm{Pa}}
\newcommand{\sibs}{\mathcal{S}}

\newcommand{\lp}{\left (}
\newcommand{\rp}{\right )}
\newcommand{\lbr}{\left [}
\newcommand{\rbr}{\right ]}
\newcommand{\lcbr}{\left \{}
\newcommand{\rcbr}{\right \}}

\usepackage[english]{babel}

\jyear{2021}%

\theoremstyle{thmstyleone}%
\theoremstyle{thmstyletwo}%

\theoremstyle{thmstylethree}%
\newtheorem{definition}{Definition}%

\newcommand{\bhmc}{{{BHMC}}}
\newcommand{\rbhmc}{{{RBHMC}}}

\algrenewcommand{\algorithmiccomment}[1]{\bgroup\hfill\footnotesize//~#1\egroup}

\begin{document}

\title[Posterior Regularized BHMC]{Posterior Regularization on Bayesian Hierarchical Mixture Clustering}


\author*[1,3]{\fnm{Weipeng Fuzzy} \sur{Huang}}\email{fuzzyhuang@tencent.com}

\author[2]{\fnm{Tin Lok James} \sur{Ng}}\email{ngja@tcd.ie}

\author[3]{\fnm{Nishma} \sur{Laitonjam}}\email{nishma.laitonjam@insight-centre.org}

\author[3]{\fnm{Neil J.} \sur{Hurley}}\email{neil.hurley@insight-centre.org}

\affil*[1]{\orgdiv{SBD Lab}, \orgname{Tencent}, \orgaddress{\city{Shenzhen}, \country{P.R. China}}}

\affil[2]{\orgdiv{School of Computer Science and Statistics}, \orgname{Trinity College Dublin}, \orgaddress{\city{Dublin}, \country{Ireland}}}

\affil[3]{\orgdiv{Insight Centre for Data Analytics}, \orgname{University College Dublin}, \orgaddress{\city{Dublin}, \country{Ireland}}}


\abstract{

Bayesian hierarchical mixture clustering (BHMC) improves traditional Bayesian hierarchical clustering by replacing conventional Gaussian-to-Gaussian kernels with a Hierarchical Dirichlet Process Mixture Model (HDPMM) for parent-to-child diffusion in the generative process.
However, BHMC may produce trees with high nodal variance, indicating weak separation between nodes at higher levels. To address this issue, we employ Posterior Regularization, which imposes max-margin constraints on nodes at every level to enhance cluster separation.
We illustrate how to apply PR to BHMC and demonstrate its effectiveness in improving the BHMC model.

}

\keywords{Posterior Regularisation, Bayesian Nonparametrics, Bayesian Hierarchical Clustering, Bayesian Algorithms}



\maketitle
\section{Introduction}
\label{sec:intro}
%

Hierarchical clustering (HC) is a well-known problem that aims to arrange data items into hierarchical structures.
The primary objective of hierarchical cluster analysis is to construct a ``meaningful structure'' for the chosen items that can either analyze the underlying relationships between the items (e.g., gene analysis) or aid in improving item retrieval (e.g., product catalogs on online retail websites).
In this context, we concentrate on the latter and strive for a structure that clusters similar items together and organizes the clusters hierarchically, allowing items in a child cluster to be viewed as variations of those in a parent cluster.


We study this problem through a Bayesian approach.
Our previous work~\citep{Huang2021inferring} proposes the \emph{Bayesian Hierarchical Mixture Clustering}~(\bhmc{}) model, which is in the same framework as other Bayesian generative approaches~\citep{williams2000mcmc,adams2010tree,neal2003density,Knowles2015}.
The generative model consists of two components: ($i$) a random walk process from a root node to a leaf node; and ($ii$) a parent-to-child random diffusion that connects parameters in a parent node to those in a child node. The common choices for the first component are nested Chinese restaurant process (nCRP)~\citep{blei2010nested,adams2010tree,Huang2021inferring}, Dirichlet Diffusion Tree~\citep{neal2003density}, and Pitman-Yor Diffusion Tree~\citep{Knowles2015}, which are asymptotically equivalent~\citep{Knowles2015}.
For the second component, a Gaussian-to-Gaussian (G2G) kernel is commonly applied~\citep{adams2010tree,neal2003density,Knowles2015}, but the \bhmc{} uses a Hierarchical Dirichlet Process Mixture Model (HDPMM) instead, making each node associated with a mixture model whose components become sparser on descent along a path in the hierarchy.

In this paper, our contribution is the development of a novel approach that incorporates Posterior Regularization (PR)~\citep{Graca2009,Ganchev2010,zhu2011infinite} to enhance the separation of sibling clusters at every level of the hierarchy.
We introduce \emph{posterior-regularized \bhmc{}} or \rbhmc{}, which utilizes the PR framework to handle Bayesian models with additional constraints.
The framework is founded on the variational inference (VI)~\citep{blei2017variational} of minimizing the Kullback-Leibler (KL) divergence between a proposed distribution and the posterior  in a constrained space.


A well-structured hierarchy should have clear separation, in particular at higher levels.
In the case of BHMC, this means that mixture components associated with each node should be close together geometrically, relative to their distance from components associated with other nodes on the same level.
However, the inference procedure can get stuck in local modes where high-level components are not sufficiently coherent.
We attribute this to the fact that the data is only generated at the leaf nodes, causing the inference to take a long time to influence higher-level mixture components.
To address this, we regularize the model to focus the optimization on distributions that exhibit the desired separation among nodes at the same level.
PR is employed to impose explicit data dependence on the choice of mixture components at every level, pushing the direct influence of data up to all levels along every path in the hierarchy.
We apply the max-margin property to nodes at every level, ensuring that the decision boundary separating the correct node for each observation is maximally distant from sibling nodes.
\par
The rest of the paper is structured as follows. Section \ref{sec:model} presents the background on Bayesian hierarchical mixture clustering and its limitations. Section \ref{sec:regbayes} presents the posterior regularized solution to address the weaknesses of Bayesian hierarchical mixture clustering, and the statistical inference of the proposed model is presented in Section \ref{sec:inference}. We conduct experimental studies to study the behavior of the proposed model.

\section{Preliminaries}
\label{sec:model}

A hierarchy, also referred to as a tree, is a data structure that manages a set of observations $\data$ with $N$ elements $\{\x_n\}_{n=1}^N$\footnote{In the rest of the paper, the term tree and hierarchy will be used interchangeably}.
Let us denote the tree by $\tree = (Z_{\tree}, E_{\tree})$ where $Z_{\tree}$ is the set of nodes and $E_{\tree}$ is the edge set in the tree.
An edge $(z, z') \in E_{\tree}$ indicates that $z$ is the parent of $z'$. The set of non-leaf nodes is denoted by $I_{\tree}$, and the height of the tree is denoted by $L$.
The set of all maximal paths is denoted by $\pathset$ where
\[
\pathset = \{\bm{z} = \langle z_0, z_1, \ldots, z_{L}\rangle \mid (z_{\ell-1}, z_{\ell}) \in E_{\tree}, \ell=1, \ldots, L \} \,
\]
where $z_0$ is the root node.
The parent of a node $z$ is denoted by $\parent(z)$, and the set of siblings of $z$ is denoted by $\sibs(z)$, where $\sibs(z) = \{z' \ne z \mid (\parent(z), z') \in E_{\tree} \}$.

\paragraph{Dirichlet process}
The BHMC relies heavily on the Dirichlet Process (DP)~\citep{ferguson1973bayesian,antoniak1974mixtures}.
The Chinese Restaurant Process (CRP) and stick-breaking process are the two well-known forms of DP.
In the CRP formulation, the $n$th customer (observation) selects a table labeled as $c_n$ when coming into a restaurant.
The randomness of $\crp(\alpha)$ is defined as
\begin{align*}
p(c_n = k \mid c_{1:n-1}, \alpha) =
\begin{cases}
{\frac{N_k}{n + \alpha}} & k \mbox{~exists} \\
{\frac{\alpha}{n + \alpha}} & k \mbox{~is~new}
\end{cases}
\end{align*}
which means that the customer selects a table proportional to the number of customers ($N_k$) sitting at the table $k$ while there is still a possibility to start a new table determined by the parameter $\alpha$. 

The stick-breaking process simulates the process of infinitely taking portions out from a remaining stick.
Assuming that the original stick is of length $1$, the process can be represented as below.
\begin{align*}
\beta_1 = \beta'_1 \qquad \beta_k = \beta'_k \prod_{i=1}^{k-1} (1-\beta'_i) \qquad \beta'_i \sim \Beta(1, \alpha) \
\end{align*}
The two representations are equivalent since $p(c_n = k \mid \alpha) = \beta_k$. $\bm{\beta} =(\beta_i)_i, i=1,2,3$ is said to follow the $\Gem(\alpha)$ distribution.

Finally, let $H$ be a base measure on a sample space and $\alpha$ a real positive concentration parameter.
We write $G \sim \DP(\alpha, H)$ if
 \begin{align*}
G = \sum_{k=1}^{\infty} \beta_k \delta_{\theta_k}
\qquad
\{\theta_k\}_{k=1}^\infty \sim H
\qquad
\bm{\beta} \sim \Gem(\alpha)\,
\end{align*}
where $\delta$ is the Dirac Delta function.
However, in a finite setting, the following approximation is used for the distribution of $\bm{\beta}$: 
$$\bm{\beta} \sim \Dir(\alpha/K, \dots, \alpha/K)$$ for a fixed large $K$, where $\Dir(\cdot,\cdot,\cdot)$ denotes the Dirichlet distribution. 

\paragraph{Nested Chinese restaurant process}
The nested Chinese Restaurant Process (nCRP) extends the CRP to a hierarchical random process.
Fixing a hierarchy height $L$, the process recursively applies the CRP to randomly traverse to the next level.

\paragraph{Hierarchical Dirichlet process mixture model}
The BHMC considers a hierarchy in which each node maintains a global book of mixture components, while keeping local weights for the components which can be completely different from each other.
This can be modeled using a \emph{hierarchical} DP (HDP)~\citep{Teh2006hier}.

If $G$ is drawn from a DP with base measure $G_0$ such that $G \sim \DP(\gamma, G_0)$ and $G_0 \sim \DP(\gamma_0, H)$, $G$ is said to be sampled from an HDP.
The 1-level HDP mixture model (HDPMM) is the limiting distribution ($K\to\infty$) of the following finite mixture model for generating observations $\x_n, n=1,2,\ldots$:
\begin{align*}
\beta_{1} \ldots \beta_K \mid \gamma_0 &\sim \Dir(\gamma_0/K,\dots, \gamma_0/K) \\
\tilde{\beta}_{1} \ldots \tilde{\beta}_{K} \mid \gamma, \bm{{\beta}} &\sim \Dir(\gamma \beta_1,\dots,\gamma \beta_K) \\
\theta_1 \ldots \theta_K \mid H & \sim H \\
c_n \mid \tilde{\bm{\beta}} &\sim \discrete(\tilde{\bm{\beta}}) \\
\x_n \mid c_n, \theta_1 \ldots \theta_K &\sim F(\theta_{c_n})
\end{align*}
where $F(\theta)$ is a distribution parameterized by $\theta$. The 1-level HDPMM which can be extended to the multilevel case.

\subsection{Bayesian hierarchical mixture clustering}
\begin{figure}[!h]
\centering
\subfloat[Specialization Relationship along a Hierarchical Path \label{fig:subsetnodes}]{
\includegraphics[width=0.35\textwidth]{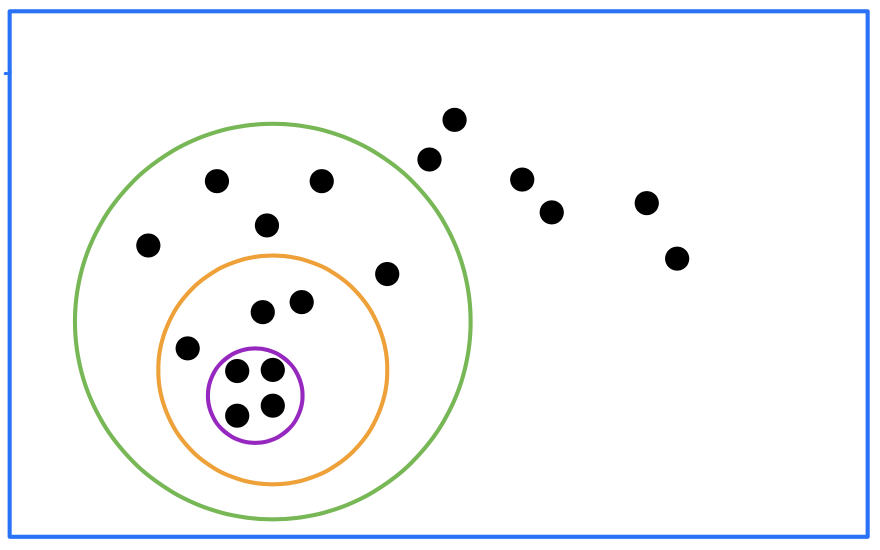}
}
~
\subfloat[Concept drift with G2G kernel \label{fig:g2gdrift}]{
\includegraphics[width=0.35\textwidth]{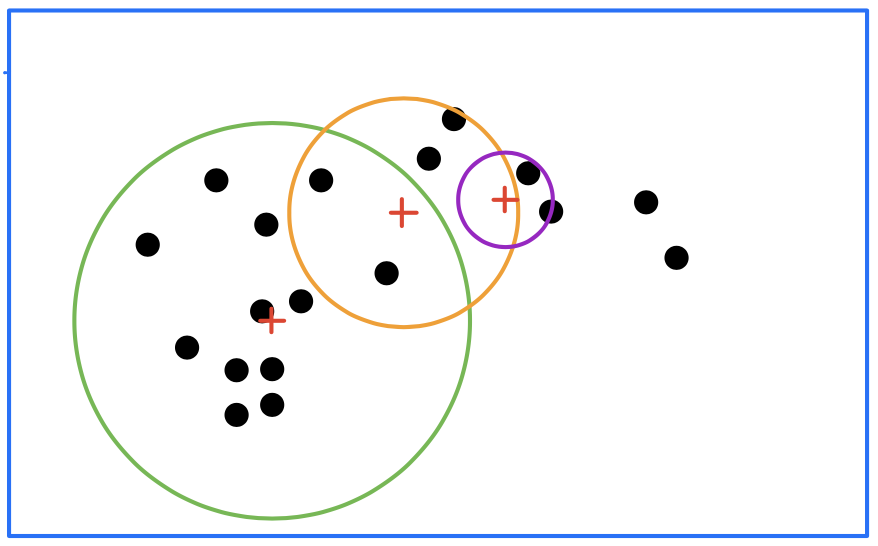}
}
\caption{G2G Bayesian hierarchical clustering}
\label{fg:kernel-comparison}
\end{figure}

\begin{figure}[!h]
\centering
\includegraphics[width=0.58\textwidth]{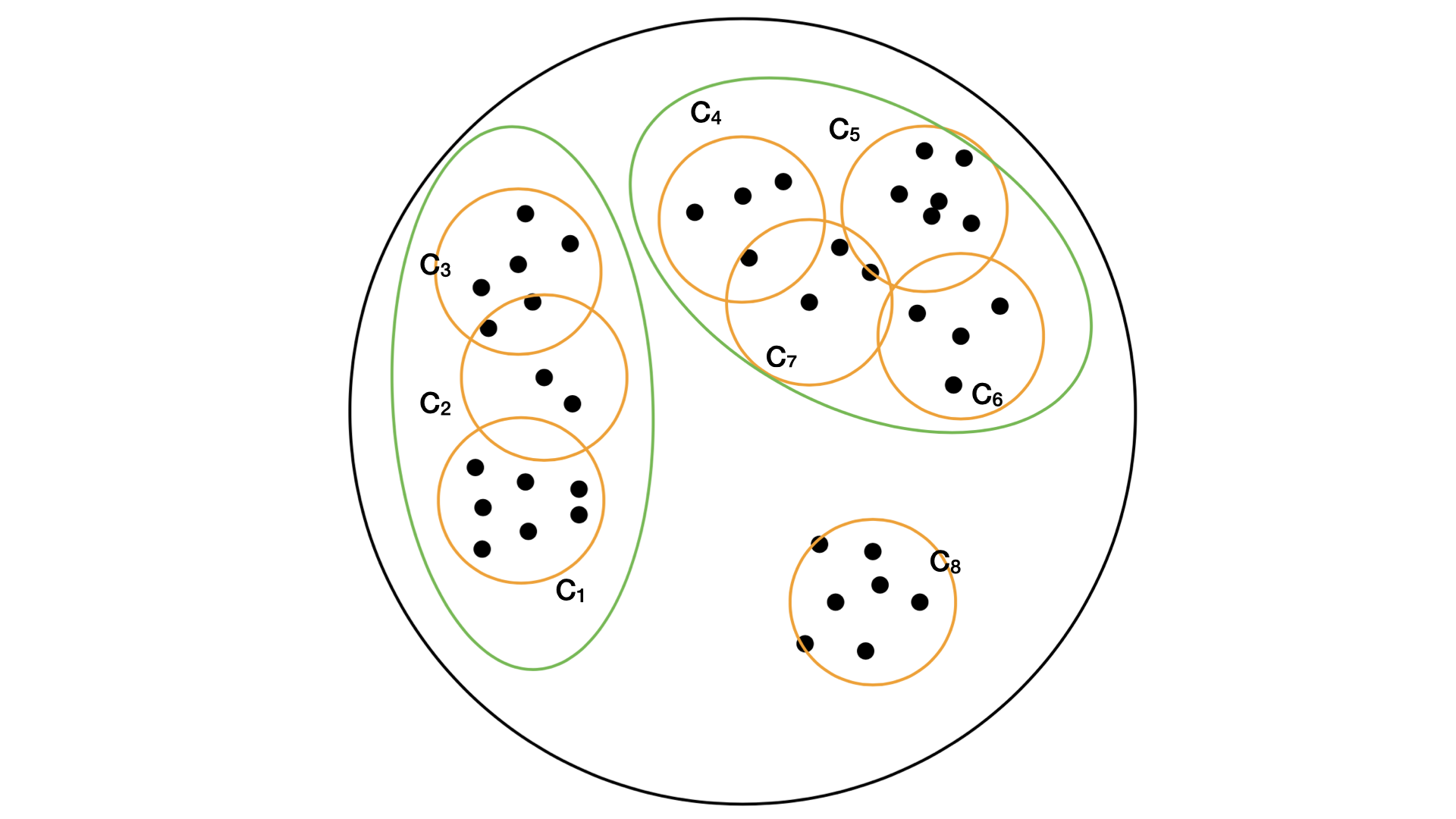}
\caption{Hierarchical Gaussian mixtures} \label{fig:hiermix}
\end{figure}

For clustering of data points located in real space, the parent-to-child transition  that has been proposed in the state-of-art is a G2G transition in which a Gaussian distribution of some mean and covariance is associated with a parent cluster and the child distribution is also Gaussian with mean drawn from the parent distribution and the variance of the distributions diminishing down the tree.
However, this model may experience ``concept drift'' along a path, as shown in~\Cref{fig:g2gdrift}, where data points at the leaf of a path may be extreme outliers of the node distribution at the start of the path.
Instead, bhmc{} uses a HDPMM to connect nodes and assumes data is from Gaussian mixture model. Child nodes have a more specialized mixture than their parent.
For instance, in~\cref{fig:hiermix}, data is represented as a Gaussian mixture of 8 clusters at the finest granularity level.
Clusters can be naturally grouped together at higher levels.
For example, data generated from one parent is a mixture of Gaussians associated with clusters $\{c_1, c_2, c_3\}$.
Its child nodes are associated with data generated from each cluster.

The \bhmc{} generates data with an nCRP having $L$ steps from the root node.
When a new node is created, local weights are sampled from the parent node through an HDP and associated with the new node.
After walking $L$ times down the hierarchy, a leaf node is reached and the observation $\x_n$ is sampled using the local weights of the leaf node.

\begin{figure}[!ht]
\centering
\begin{forest}
  for tree={
    fit=band,
  }
  [\scalebox{0.13}{\includegraphics{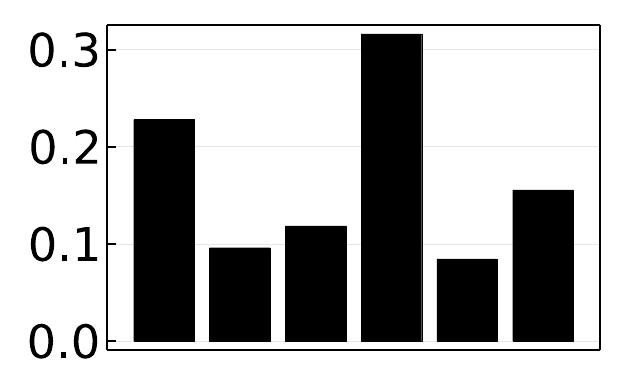}} $z_0$
    [\scalebox{0.13}{\includegraphics{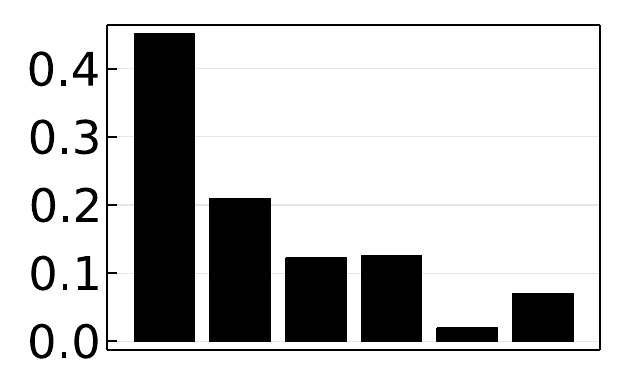}} $z_1$
      [\scalebox{0.13}{\includegraphics{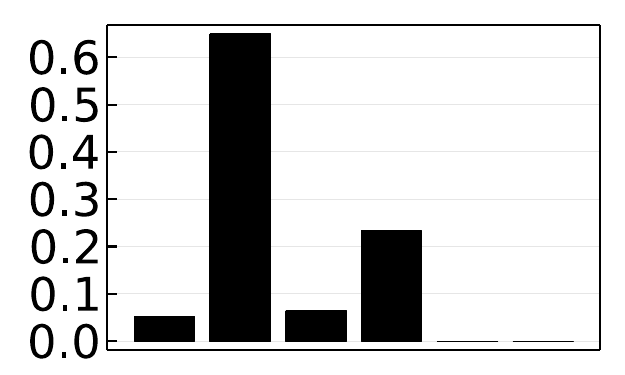}} $z_4$]
      [\scalebox{0.13}{\includegraphics{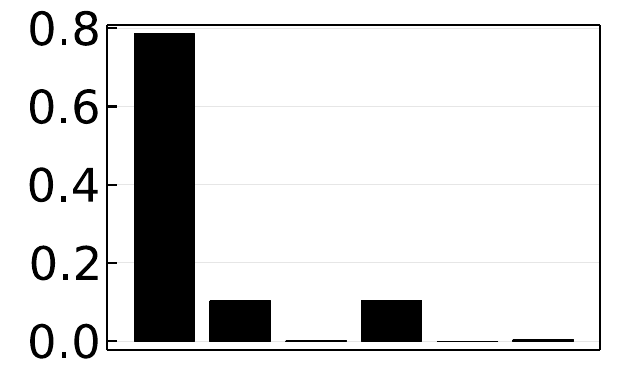}} $z_5$]
    ]
    [\scalebox{0.13}{\includegraphics{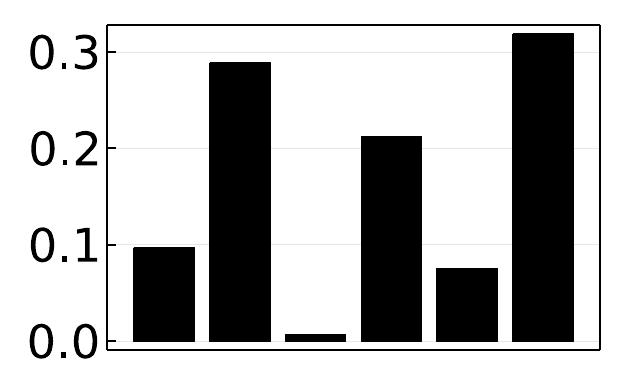}} $z_2$
      [\scalebox{0.13}{\includegraphics{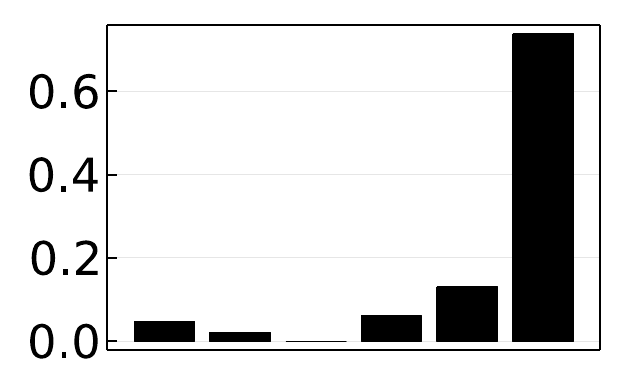}} $z_6$]
      [\scalebox{0.13}{\includegraphics{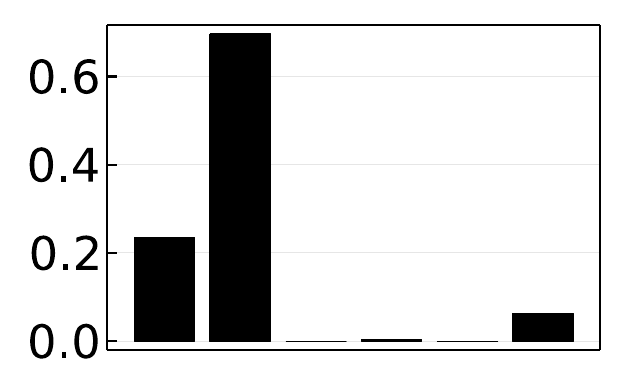}} $z_7$]
    ]
    [\scalebox{0.13}{\includegraphics{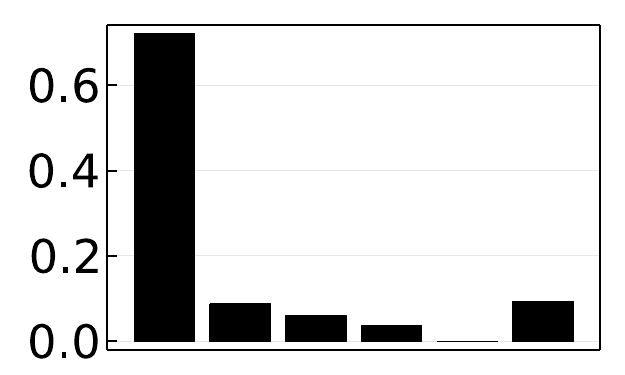}} $z_3$
      [\scalebox{0.13}{\includegraphics{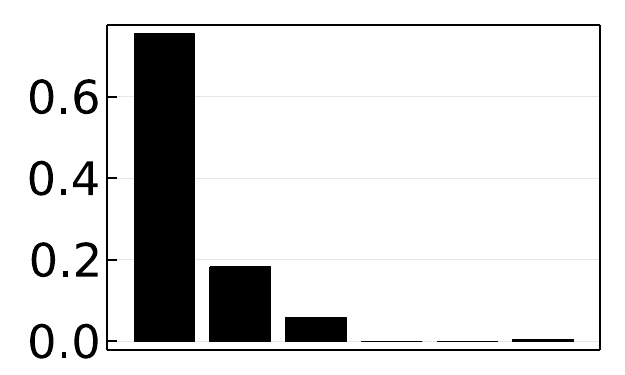}} $z_8$]
    ]
  ]
\end{forest}
\caption{An example of the BHMC generative process where each plot indicates the mixing proportion of the corresponding node}
\label{fg:bhmc-example}
\end{figure}

To better illustrate the generative process, we resort to the finite setting of the DP in the upcoming example.
The generative example in~\Cref{fg:bhmc-example} fixes $K=6$ and $L=2$, i.e., there are $6$ global components and the height of the tree is $2$.
First, a path through the hierarchy is a sequence, e.g., $\bm{z} = \langle z_0, z_1, z_4 \rangle$.
Each observation $\x_n$ is assigned a path indicator variable, which we denote by $\bm{v}_n$.
Specifically, if observation $n$ traverses along path $\bm{z} \in \pathset$, the assignment follows $\bm{v}_n = \bm{z}$.
Further, we write ${v}_{n \ell}$ to point to the node at the $\ell$th level of the path.
Mixing proportions associated with a certain node $z$, $\bm{\beta}_z$, along a path are connected via a multilevel HDP.

Upon arriving at a leaf node $v_{n L}$, the $n$th observation is assigned a component label $c_n$ depending on the mixing proportions $\bm{\beta}_{v_{nL}}$  such that $c_n \sim \D(\bm{\beta}_{v_{n L}})$.
Eventually, the process samples $\x_n \sim F(\theta_{c_n})$ where $\theta$ is sampled from the base measure $H$.
The corresponding process for observation $n$ is summarized as
\begin{align*}
&\bm{v}_n  \sim \mbox{nCRP}(\alpha) \\
&\bm{\beta}_{z_0} \sim \Dir\lp \gamma_0/K, \ldots, \gamma_0/K \rp \qquad
\bm{\beta}_{v_{n 1}} \sim \Dir(\gamma \bm{\beta}_{z_0}) \qquad \bm{\beta}_{v_{n 2}} \sim \Dir(\gamma \bm{\beta}_{v_{n 1}}) \\
&\x_n \sim F(\theta_{c_n}) \qquad c_n \sim \discrete(\bm{\beta}_{v_{n 2}}) \qquad \theta_1, \ldots, \theta_K \sim H
\, .
\end{align*}
\Cref{fg:bhmc-example} provides an example of how mixing proportions can propagate through levels.
In practice, this can result in sparser components at lower levels.
If the probability of a mixture component becomes negligible at any node in the tree, it is highly likely to remain negligible for all descendant nodes.

\begin{algorithm}[ht]
\caption{\sc Generative Process of the BHMC (Infinite Setting)}
\label{alg:generative}
\begin{algorithmic}[1]
\State Sample $\beta_{z_0 1}, \dots, \beta_{z_0 K} \sim \Gem(\gamma_0)$ 
\State Sample $\theta_1, \dots, \theta_K \sim H$
\For{$i=1 \ldots N$}
  \State $v_{n 0} \gets z_0$
  \For{$\ell = 1 \ldots L$}
    \State Sample $v_{n \ell}$ by CRP($\alpha$)
    \State $z, z' \gets v_{n (\ell-1)}, v_{n \ell}$
    \If{$z'$ is new}
      \State Sample $\bm{\beta}_{z'} \sim \DP\left(\gamma, \bm{\beta}_{z}\right)$
      \State Attach $(z, z')$ to the tree $\tree$
    \EndIf
  \EndFor
  \State Sample $c_n \sim \discrete(\bm{\beta}_{v_{n L}})$
  \State Sample $x_n \sim F(\theta_{c_n})$
\EndFor
\end{algorithmic}
\end{algorithm}
\Cref{alg:generative} demonstrates a detailed generative process
for the BHMC.
Let us denote the variables in the model by $\model_0$ such that $\model_0 = \{\m{o}, \tilde{\m{B}}, \m{V}, \m{c}, \bm{\theta} \}$.
The variables are $\tilde{\m{B}} = \{\bm{\beta}_z\}_{z \in \nodeset \setminus \{z_0\}}$, $\m{o} = \{o_k \}_{k=1}^{\infty}$, $\m{V} = \{\bm{v}_n \}_{n=1}^N$, $\m{c} = \{c_k \}_{k=1}^{\infty}$, and $\bm{\theta} = \{ \theta_k \}_{k=1}^{\infty}$.
Under an infinite setting, we sample $\bm{\beta}_{z_0} \sim \Gem(\gamma_0)$ for the root node $z_0$.
That is, considering $o \sim \Beta(1, \alpha)$ for any $o$, we write $\beta_{z_0 k} = o_k \prod_{k'=1}^{k-1} (1 - o_{k'})$ and $\beta_{z_0}^* = 1 - \sum_{k=1}^K \beta_{z_0 k}$ where $K$ represents a finite integer.
It follows that $p(\bm{\beta}_{z_0} \mid \gamma_0) \equiv p(\m{o} \mid \gamma_0)$.
For any $(z, z') \in E_{\tree}$, the model defines $\bm{\beta}_{z'} \sim \Dir(\gamma \bm{\beta}_z)$.
We write $\tilde{\m{B}}$ to maintain the mixing proportions for the non-root nodes, and the total mixing proportions $\m{B} = \tilde{\m{B}} \cup \{\bm{\beta}_{z_0}\}$.
Thus we obtain $p(\m{B} \mid \gamma, \gamma_0) \equiv p(\m{o} \mid \gamma_0) p(\tilde{\m{B}} \mid \gamma)$.
The model is summarized by~\Cref{eq:model-prob}.
\begin{align}
\label{eq:model-prob}
&p(\m{V}, \m{c}, \m{B}, \bm{\theta} \mid \m{X}, \alpha, \gamma, \gamma_0, H, L) \nonumber \\
&\propto
p(\data \mid \m{V}, \m{c}, \bm{\theta}, \m{B}) p(\m{c} \mid \m{V}, \m{B}) p(\m{V} \mid  \alpha, L) p(\bm{\theta} \mid H) p(\m{B} \mid \gamma_0, \gamma) \nonumber \\
&= \prod_n p(\x_n \mid c_n, \bm{\theta})\prod_n p(c_n \mid \bm{v}_n, \bm{\beta}_{v_{n L}}) \prod_n p(\bm{v}_n \mid \alpha, L)  \nonumber \\
&\quad \prod_k p(\theta_k \mid H) \lbr p(\bm{\beta}_{z_0} \mid \gamma_0) \prod_{(z, z') \in \edgeset} p(\bm{\beta}_{z'} \mid \gamma, \bm{\beta}_z) \rbr \,.
\end{align}


\subsubsection{Weaknesses of the BHMC}

\begin{figure}[!ht]
\centering
\includegraphics[width=0.6\textwidth]{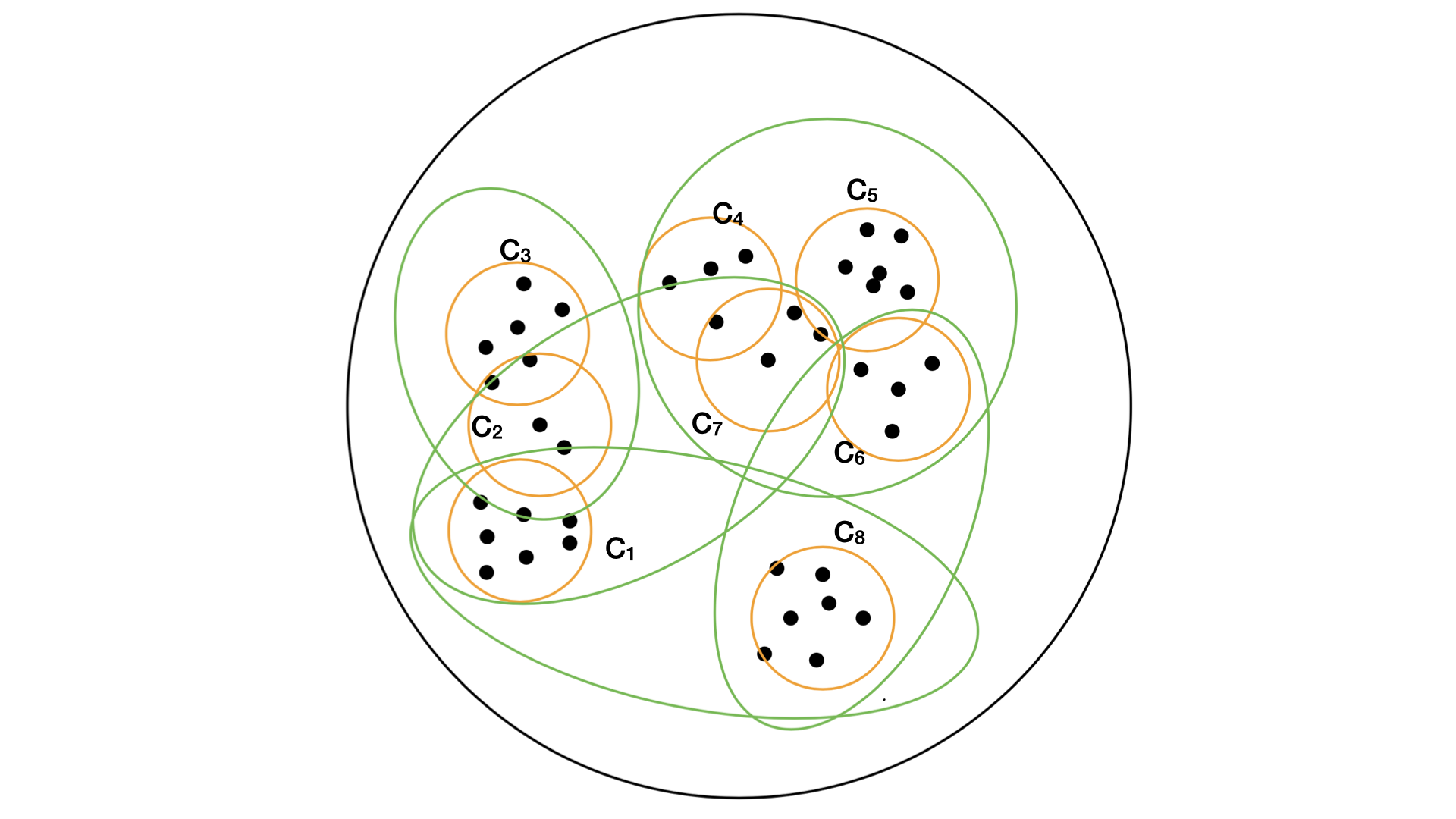}
\caption{Hierarchical clustering with shared components.} \label{fig:overlapclusters}
\end{figure}

The study in~\citet{Huang2021inferring} shows that the BHMC performs well compared to other HC algorithms (either Bayesian or Non-Bayesian) based on commonly used evaluation metrics.
However, a weakness of the model is that it allows sibling nodes under a common parent to have shared components, which can result in poor separation and make it challenging to identify the correct path to a target item.
A more human understandable hierarchy can be achieved by ensuring reasonably good separation, resulting in better controlled nodal variance for any level.
The BHMC lacks a mechanism to ensure well-separated sibling nodes in the hierarchy.
An extreme example of a valid hierarchical clustering generated from the BHMC is illustrated in~\Cref{fig:overlapclusters}.

Taking inspiration from work on posterior regularization (PR)~\citep{Graca2009,Ganchev2010,Zhu2014regbayes}, we aim to overcome the limitation of the BHMC model by using PR to enforce sibling separation during Bayesian learning.
As noted in~\citet{Graca2009}, incorporating even minor additional constraints into generative models can be highly challenging for the posterior inference.
However, by utilizing a posterior constraint, we can guide inference towards solutions that exhibit the desired separation without altering the generative process.

\section{Posterior Regularized Solution}
\label{sec:regbayes}
PR was developed upon VI for learning Bayesian models with constraints~\citep{Zhu2014regbayes}.
The objective of VI is to minimize the KL-divergence (denoted by $\KL$) between a proposed distribution and the posterior distribution.
The PR framework includes a penalty function based on a set of feature constraints to the objective.
We use $\model$ to denote an arbitrary model and $\E_q$ to denote the expectation over a distribution $q$.

\begin{definition}[RegBayes]
Let $q(\model)$ be a distribution proposed to approximate the posterior $p(\model \mid \data)$.
The PR method is to solve
\begin{align}
\label{eq:kl-def}
\min_{q(\model), \bm{\xi}} &~ \KL\infdivx{q(\model)}{p(\model, \data)} + U(\bm{\xi})\quad s.t. ~ q(\model) \in \mathbb{P}_{cs}(\bm{\xi})
\end{align}
where $p(\model, \data)$ denotes the joint probability of the model $\model$ and the data $\data$, and $U(\bm{\xi})$ is a regularization function obtained upon the constraint space $\mathbb{P}_{cs}(\bm{\xi})$.
\end{definition}
The constrained space $\mathbb{P}_{cs}(\bm{\xi})$ is parameterized by $\bm{\xi}$ which itself enters the objective function. In contrast to standard VI which performs unconstrained optimization over $q(\model)$, the regularized Bayes solution differs in two ways: (1) the optimization over $q(\model)$ is performed in the constrained space $\mathbb{P}_{cs}(\bm{\xi})$, and (2) it performs joint optimization over $q(\model)$ and $\bm{\xi}$.

The conventional approach involves transforming the objective into its dual form.
The model variables are learned using VI, and the dual variables are solved using optimization tools.
However, the VI approach may result in a large number of dual variables and thus intractability in our model.
Fortunately, we can use the Markov Chain Monte Carlo (MCMC) method to infer the proposed model.

\subsection{RBHMC}
\label{sec:motivation}
We aim to restrict the search space to solutions where the separation between the siblings is reasonably large.
To this end, we appeal to max-margin, which is used in e.g. support vector machine (SVM) classification to ensure adequate separation between classes.
To clarify its use in HC, it is beneficial to revisit its formulation in this context.

Starting from the simplest SVM binary classification, we consider that a observation $\x_n \in \data$ could be assigned with a label $\tau_n$ where $\tau_n \in \{-1, 1\}$.
Now, we augment it with a latent discriminant variable $\bm{\eta}$ so that
\begin{align*}
  \tau_n \lp \bm{\eta}^\top \x_n + \eta_0 \rp \ge \varepsilon / 2
  \implies
  \begin{cases}
    \lp \bm{\eta}^{\top} \x_n + \eta_0 \rp \ge \varepsilon / 2  & \tau_n = 1 \\
    -\lp \bm{\eta}^{\top} \x_n + \eta_0 \rp \ge \varepsilon / 2  & \tau_n = -1
  \end{cases} \, .
\end{align*}
Thus, $\bm{\eta}^\top \x + \eta_0 = 0$ is the hyperplane boundary that separates the two classes, and the data on the two sides are at least $\varepsilon/2$ units away from the hyperplane~\citep{hastie2009elements}.
However, the data may not be linearly separable, so that no such boundary can be found. 
Thus, a non-negative slack variable $\xi_n \ge 0$ is introduced and we write
\begin{align*}
  \tau_n \lp \bm{\eta}^\top \x_n + \eta_0 \rp \ge \varepsilon / 2 - \xi_n / 2 \, .
\end{align*}
Hence, $\sum_n \xi_n$ is thought of as the sum of the distances by which the data lie on the wrong side of their margin defined by $\varepsilon$.

However, modifications are required for handling multiple classes.
One method is to associate a different $\bm{\eta}$ with each cluster.
This $\bm{\eta}$ can be employed to form a boundary to separate the data in this cluster from the rest, namely, the one-vs-rest approach~\citep{shalev2014understanding}.
Considering a class denoted with $z$ or $z'$, where $z$ is the node that observation $n$ belongs to, the optimization will have the following constraints:
\begin{multline}
\label{eq:constraint-single-lvl}
\begin{cases}
  \lp \bm{\eta}_{z}^{\top} \x_n + \eta_0 \rp \ge \varepsilon / 2 - \xi_n / 2 & \mbox{observation $n$ belongs to node $z$} \\
  -\lp \bm{\eta}_{z'}^{\top} \x_n + \eta_0 \rp \ge \varepsilon / 2 - \xi_n / 2 & \forall z': z' \ne z
\end{cases} \\
\implies
\mathbb{P}_{cs}^{(n)} = \lcbr (\bm{\eta}_{z} - \bm{\eta}_{z'})^\top \x_n \ge \varepsilon - \xi_n ~\big |~ \forall z': z' \ne z, \xi_n \ge 0 \rcbr
\end{multline}
where $\mathbb{P}_{cs}^{(n)}$ denotes the constraint space for observation $n$.
This work extends this idea to a multilevel case to fit the task of HC, where the constraint can be applied to clusters under the same parent.
Given the constraint space, we can apply the PR to restrict the search space for the Bayesian inference.


\subsection{Specifications of the RBHMC}
Let us specify the detail of the PR model in this section.
The variable set for the \bhmc{} is denoted by $\model_0$ and the whole model is denoted by $\model = \model_0 \cup \{ \bm{\eta} \}$ where $\bm{\eta}$ refers to a set of discriminant variables as discussed in~\Cref{sec:motivation}.
Let us extend~\Cref{eq:constraint-single-lvl} to a constraint space for HC.
We apply~\Cref{eq:constraint-single-lvl} for the assignment of the observation $n$ at each level $1 \le \ell \le L$ rather than only at the leaf level.
These constraints then ensure that the hierarchy is well separated under each internal parent node.
Recalling that $v_{n \ell}$ represents the correct node for observation $n$ at level $\ell$ and $\sibs(v_{n \ell})$ represents all siblings of $v_{n \ell}$, the constraint space may be written as
\begin{align}
\label{eq:constrained-space}
{\mathbb P}_{cs}(\bm{\xi}) =
\left\{
q( \model ) ~\Bigg |~
\begin{array}{l}
  \forall n,~
  \forall \ell,~
  \displaystyle \forall z \in \sibs(v_{n \ell}): \\
  \E_q [\bm{\eta}_{v_{n \ell}} - \bm{\eta}_{z}]^{\top} \x_n \ge \varepsilon_{n \ell}^{\Delta} - \xi_n,~\xi_n \ge 0
\end{array}
\right\}
\end{align}
where $\varepsilon_{n \ell}^\Delta$ is the cost of choosing the node $z$ over the assigned child $v_{n \ell}$. That is, $\varepsilon_{n \ell}^\Delta = \varepsilon_0 \mathbbm{1}(z \ne v_{n \ell})$ and $\varepsilon_0$ can be regarded as a hyperparameter.
Denoting $C$ the scale factor, we define the penalty function $U(\bm{\xi}) = 2C \sum_n \xi_n$.
Therefore, based on~\Cref{eq:kl-def,eq:constrained-space}, the objective can be detailed as
\begin{subequations}
\label{eq:constraint-obj}
\begin{align}
  \min_{q(\model), \bm{\xi}} \quad & \KL\infdivx{q(\model)}{p(\model, \data)} + 2 C \sum_n \xi_n \\
  s.t. \quad & \forall n, \forall \ell, \forall z \in \sibs(v_{n \ell}):
  \xi_n \ge \varepsilon_{n \ell}^{\Delta} - \E_q [\bm{\eta}_{v_{n \ell}} - \bm{\eta}_{z}]^{\top} \x_n , \xi_n \ge 0 \label{eq:constraint-obj-v1} \, .
\end{align}
\end{subequations}
Taking into account the above objective for fixed $q(\model)$, such that $\E_q [\bm{\eta}_{v_{n \ell}} - \bm{\eta}_{z}]^{\top} \x_n$ is fixed, it may be observed that the optimal $\xi_n$ is obtained at $\xi_n = 0$ when $\forall \ell, \forall z \in \sibs(v_{n \ell}): \varepsilon_{n \ell}^{\Delta} - \E_q [\bm{\eta}_{v_{n \ell}} - \bm{\eta}_{z}]^{\top} \x_n <0$ and otherwise $\xi_n = \max_{\ell, z \in \sibs(v_{n \ell})}(\varepsilon_{n \ell}^{\Delta} - \E_q [\bm{\eta}_{v_{n \ell}} - \bm{\eta}_{z}]^{\top} \x_n)$. 
That is, the optimal $\xi_n$ satisfies~\citep{shalev2014understanding}
\begin{align}
  \label{eq:hinge-loss}
  \forall n: \xi_n = \max \lp 0, \max_{\ell, z \in \sibs(v_{n \ell})} \varepsilon_{n \ell}^{\Delta} - \E_q [\bm{\eta}_{v_{n \ell}} - \bm{\eta}_{z}]^{\top} \x_n \rp \,.
\end{align}

To further simplify the notation, let us introduce an auxiliary variable:
\begin{align}
  \label{eq:auxiliary_rho}
  \rho_{n \ell z} \coloneqq \varepsilon_{n \ell}^\Delta - (\bm{\eta}_{z} - \bm{\eta}_{n \ell})^{\top} \x_n
\end{align}
We emphasize that $\rho_{n \ell z}$ is not a new variable but introduced for simplifying the notation.
\Cref{eq:constraint-obj} can thus be equivalently expressed as
\begin{align}
\label{eq:raw-obj}
\min_{q(\model)}~ \KL\infdivx{q(\model)}{p(\model, \data)} +  2 C \sum_n \max \left(0, \max_{\ell, z \in \sibs(v_{n \ell})} \E_q \left[ \rho_{n \ell z} \right] \right)\,.
 \end{align}
This is the standard technique for SVM to transform the constrained problem to be unconstrained~\citep{shalev2014understanding}.

Solving~\Cref{eq:raw-obj} is non-trivial due to the regularization term.
An alternative is to minimize an upper bound of the term as a surrogate.
To derive the bound, we first note that the maximum function $\max(\bm{a})$ where $\bm{a} \in \mathbb{R}^m$ is known to be convex.
In addition, given a convex function $h:\mathbb{R}^m \mapsto \mathbb{R}$, $\max(const., h(\bm{a}))$ is also convex on $\bm{a}$~\citep{boyd2004convex}.
Therefore, we obtain
$$
\max \left(0, \max_{\ell, z \in \sibs(v_{n \ell})} \E_q \lbr \rho_{n \ell z} \rbr \right)
\le
\E_q \lbr \max \left(0, \max_{\ell, z \in \sibs(v_{n \ell})} \rho_{n \ell z} \right) \rbr \,.
$$
Finally, we solve the following objective,
\begin{align}
  \label{eq:obj}
  &\min_{q(\model)}~ \KL\infdivx{q(\model)}{p(\model, \data)} +  2 C \sum_n \E_q \lbr \max \left(0, \max_{\ell, z \in \sibs(v_{n \ell})} \rho_{n \ell z} \right) \rbr\,.
\end{align}
Heuristically, the above regularization minimizes the maximal margin violation between any node in the assigned path from its siblings among all levels.

One can obtain an analytical solution to $q(\model)$ through employing the variational derivations over the objective.
For instance, \cite{chen2014robust} proved that
\begin{align*}
q^*(\model)
&=  \argmin_{q(\model)} ~ \KL\infdivx{q(\model)}{p(\model, \data)} + 2 C \sum_n \E_{q(\model)} \left[ \max \left(0, \max (\bm{\rho}) \right) \right] \\
&\propto p(\model \mid \data) \prod_n \exp\lcbr - 2C \max \lp 0, \max (\bm{\rho}) \rp \rcbr \,.
\end{align*}
The proof establishes a connection between the objective and the Euler-Lagrange equation, and subsequently derives the updates by setting the derivatives to $0$.

In our case, we specify $\bm{\rho} = \lcbr \rho_{n \ell z} \mid \forall n, \forall \ell, \forall z \in \sibs(v_{n \ell}) \rcbr$.
The optimal solution of the objective, $q^*(\model)$, is defined as
\begin{align}
\label{eq:closed-form}
q^*(\model) \propto p(\model \mid \data) \prod_n \exp\lcbr - 2C \max \lp 0, \max_{\ell, z \in \sibs(v_{n \ell})}\rho_{n \ell z} \rp \rcbr \,.
\end{align}
Finally, the process to infer $q^*(\model)$ will be derived based upon this formulation.

\section{Inference}
\label{sec:inference}
The RBHMC incorporates both the PR-imposed random variables and the original random variables from the BHMC, which were fully analyzed in~\citet{Huang2021inferring}. 
Thus, this section primarily focuses on analyzing the PR-imposed variables and their associated sampling steps. 
Next, we review the sampling details of the BHMC and connect all components to illustrate the complete inference procedure. Finally, we discuss the criterion for selecting the output hierarchy to conclude this section on algorithm implementation.

\subsection{Analyzing the PR-imposed variables}
In~\Cref{eq:closed-form}, a non-trivial regularization term 
\[
\prod_n \exp\lcbr - 2C \max \lp 0, \max_{\ell, z \in \sibs(v_{n \ell})}\rho_{n \ell z} \rp \rcbr
\] 
is introduced.
One has to decompose the term for sampling the random variables $\bm{\rho}$.
This can be achieved through data augmentation~\citep{polson2011data,chen2014robust}.

Recall the definition of $\bm{\rho}$ in~\Cref{eq:auxiliary_rho} and its usage in~\cref{eq:raw-obj}.
We define $s_{n} = (\ell, z)$, a tuple encapsulating the level $\ell$ and the sibling $z$ that maximizes $\rho_{n \ell z}$, i.e.
$s_{n} := (s_{n1}, s_{n2}) := (\ell, z) := \argmax_{\ell, z \in \sibs(v_{n \ell})} \rho_{n \ell z}$
such that $s_{n}$ is a deterministic function of the random variables $\bm{v}_n, \bm{\eta}$ and $\x_n$.
The corresponding set is written as $\m{s} = \{ s_n \}_{n=1}^N$.
Accordingly, $\rho_{n s_n}$ represents the maximum $\rho_{n \ell z}$ corresponding to that tuple.
\cite{polson2011data} demonstrated that, for any arbitrary real $\rho$,
\begin{align*}
\exp\{-2C\max(0, \rho)\}
&= \int_0^\infty \frac 1 {\sqrt{2 \pi \lambda}} \exp\left\{ - \frac {(C\rho + \lambda)^2} {2 \lambda} \right \} d \lambda \propto \int_0^\infty p(\lambda \mid \rho) d\lambda
\end{align*}
where $p(\lambda \mid \rho)$ is the density function of the Generalized Inverse Gaussian (GIG) distribution $\gig(\lambda; 1/2, 1, C^2\rho^2)$, defined as
\begin{align}
  \label{eq:gig-pdf}
  \gig(x; \kappa, a, b) \propto x^{\kappa-1} \exp \lcbr - {\lp a x + {b/x} \rp} / {2} \rcbr \,
\end{align}
where $a > 0, b > 0$.
The set of augmented variables $\bm{\lambda} = \{ \lambda_{n} \}_{n=1}^N$ hence follows
\begin{align}
\label{eq:pseudo_part}
&p(\lambda_{n} \mid \x_n, \bm{v}_n, \bm{\eta}) \nonumber \\
&\propto
\frac 1 {\sqrt{ \lambda_{n}}} \exp\left\{ - \frac {(C\rho_{n s_n} + \lambda_{n})^2} {2 \lambda_{n}} \right \} \nonumber \\
&= \frac 1 {\sqrt{ \lambda_{n}}} \exp\left\{ - \frac {[C (\varepsilon_{0} \mathbbm{1}(s_{n2} \ne v_{n s_{n1}}) - (\bm{\eta}_{v_{n s_{n1}}} - \bm{\eta}_{s_{n2}})^{\top} \x_n) + \lambda_{n}]^2} {2 \lambda_{n}} \right \} \, .
\end{align}
One can further derive that
\begin{multline}
\label{eq:augmented-obj}
q^*(\model_0, \bm{\eta}, \bm{\lambda}) \propto p(\m{B}, \bm{\theta}, \bm{\eta}) \\
\prod_n  p \lp \x_n \mid \bm{v}_n, c_n, \m{B}, \bm{\theta}, \bm{\eta} \rp p(c_n \mid \bm{\beta}_{v_{nL}}, \bm{v}_n) p(\lambda_{n} \mid \x_n, \bm{v}_n, \bm{\eta}) p(\bm{v}_n) \,.
\end{multline}
It follows that the target posterior $q^*(\model_0, \bm{\eta})$ can be approached by sampling from $q^*(\model_0, \bm{\eta}, \bm{\lambda})$ and dropping the augmented variables.

\subsection{Sampling the PR-imposed variables}
Based on the above decomposition, we specify the corresponding sampling strategy for all variables.
Note that, although sampling paths is similar to that in~\citet{Huang2021inferring}, the change of the employed likelihood function leads to sampling deviations at the detailed level.

\bmhead{Sampling $\m{V}$ and $\m{s}$}

There could be numerous divergence points to new paths.
As a result, sampling the path through a Gibbs step may be impractical, and even a slice sampling scheme~\citep{neal2003slice,walker2007sampling,Kalli2011} could suffer from intractability, especially when the hyperparameters $L$ and $\alpha$ jointly create a large initial tree.
The tree data structure is memory-intensive, even for moderately sized datasets, making it difficult to ``downsize'' the path candidate pool using the posterior slice threshold---a common strategy for slice sampling.
Determining an appropriate initial candidate pool size that ensures expected performance requires independent research, which we leave to future work.
Therefore, we opt to use the Metropolis-Hastings (MH) sampling for the path, similar to the approach detailed in~\citet{Huang2021inferring}.


In the RBHMC, the MH scheme proposes a new path $\bm{v}_n$ using nCRP, along with a deterministic $s_n$ and new $\bm{\beta}$ if the path samples new nodes.
The proposal is accepted with the acceptance probability $\mathcal{A}$.
However, MH can be combined with partially collapsed Gibbs to improve convergence efficiency, as pointed out by~\citet{van2011partially}.
Adopting the approach in~\cite{Huang2021inferring}, we consider an MH step to sample $\m{V}$ with $\m{c}$ marginalized, followed immediately by a step sampling $\m{c}$ from the complete conditional.
This order adopts~\cite{Dyk2015MH}'s paradigm to preserve the stationary distribution of the chain in an inference combining MH and partially collapsed Gibbs steps.

Let us consider a joint (regularized) likelihood $\tilde{f}(\cdot)$ of $\x_n$ and $\lambda_n$ which can be written as
\begin{align}
\label{eq:jointreglh}
\tilde{f} \lp \x_n, {\lambda}_n; \bm{v}_n, \m{B}, \bm{\theta}, \bm{\eta} \rp = p(\x_n \mid \bm{v}_n, \bm{\beta}_{v_{n L}}, \bm{\theta}) p(\lambda_{n} \mid \x_n, \bm{v}_n, \bm{\eta}) \, .
\end{align}
The first term is the marginal likelihood that collapses $\m{c}$, and the second term is given by~\Cref{eq:pseudo_part}.
A new cluster is associated with the predictive likelihood, $f^*(\x_n)$, implied by the base measure $H$, such that $f^*(\x_n) = \int f(\x_n; \theta) d p(\theta \mid H)$. Hence, summing over all $K$ existing clusters as well as the possibility of generating a  new cluster, we obtain
$
p(\x_n \mid \bm{v}_n, \bm{\theta}, \bm{\beta}_{v_{nL}})
= \sum_{k=1}^K \beta_{v_{nL} k} f(\x_n; \theta_k) + \beta_{v_{nL}}^* f^*(\x_n) \,.
$
Following~\Cref{eq:augmented-obj}, the marginal posterior $q^*_{\neg \m{c}}(\model_0, \bm{\eta}, \bm{\lambda})$ which collapses $\m{c}$, satisfies
\begin{multline}
\label{eq:augmented-obj-marginal}
q^*_{\neg \m{c}}(\model_0, \bm{\eta}, \bm{\lambda}) \propto p(\m{B}, \bm{\theta}, \bm{\eta})
\prod_n  p \lp \x_n \mid \bm{v}_n, \m{B}, \bm{\theta}, \bm{\eta} \rp p(\lambda_{n} \mid \x_n, \bm{v}_n, \bm{\eta}) p(\bm{v}_n) \,.
\end{multline}

Now, the acceptance probability of the MH scheme is
\begin{align*}
\mathcal{A} &= \min \lp 1,~ \frac {q^*_{\neg \m{c}}(\model_0', \bm{\eta}, \bm{\lambda}) \psi(\bm{v}_n, \m{B}, {s}_n \mid \bm{v}_n', \m{B}', {s}_n')} {q^*_{\neg \m{c}}(\model_0, \bm{\eta}, \bm{\lambda}) \psi(\bm{v}_n', \m{B}', {s}_n' \mid \bm{v}_n, \m{B}, {s}_n)} \rp \\
&= \min \lp1, ~ \frac {\tilde{f} \left( \x_n, \lambda_n; \bm{v}_n', \m{B}', \bm{\theta}, \bm{\eta} \right)} {\tilde{f} \left( \x_n, \lambda_n; \bm{v}_n, \m{B}, \bm{\theta}, \bm{\eta} \right)}  \rp
\end{align*}
where the prime denoted variables are those for modification given the new proposal and $\psi(\cdot)$ denotes the density of the proposal distribution.
The derivation detail is available in~\Cref{sec:rbhmc-ap-accept}.

\bmhead{Sampling $\eta$}
With some algebraic manipulation, it may be shown that the discriminant variable $\bm{\eta}_z$ can be sampled as follows:
\begin{align}
  \bm{\eta}_z \mid \data, \m{s}, \bm{\lambda}, \model \setminus \{ \bm{\eta}_z \} \sim \N \left( \m{\Lambda}_z^{-1} \bm{\nu}_z, \m{\Lambda}_z^{-1} \right) \, ,
\end{align}
where
\begin{subequations}
\begin{align}
\tilde{C}_n
&\coloneqq {C\lambda_{n}} + C^2 \varepsilon_{0} \\
\bm{\nu}_z
&= \frac 1 {\lambda_{n}} \lcbr \sum_{n: v_{n \ell}=z} \left(\tilde{C}_n + C^2 \bm{\eta}_{s_{n 2}}^{\top} \x_n \right) - \sum_{n: s_{n 2}=z} \left( \tilde{C}_n - C^2 \bm{\eta}_{v_{n \ell}}^{\top} \x_n \right) \rcbr \x_n^{\top} \\
\m{\Lambda}_z
&= \nu_0^2 \m{I} + {C^2} \sum_{n} \mathbbm{1}(v_{n \ell} = z~||~s_{n 2} = z) \frac {\x_n \x_n^{\top}} {\lambda_{n}} \, .
\end{align}
\end{subequations}
The derivation steps are provided in~\Cref{sec:rbhmc-sdv}.

\bmhead{Sampling $\lambda$}
\Cref{eq:gig-pdf} facilitates sampling $\lambda_n$ by $\lambda_n \mid \rho_{n s_n} \sim \gig(1/2, 1, C^2 \rho_{n s_n}^2)$.
However, rather than sampling $\lambda_n$ directly, one can sample the reciprocal of $\lambda_{n}$ from an Inverse Gamma (IG) distribution:
\begin{equation}
\label{eq:reci-lambda}
\lambda_{n}^{-1} \mid \rho_{n s_n}, s_n, \bm{v}_n \sim \ig \lp {\lvert C\rho_{n s_{n}} \rvert}^{-1}, 1 \rp \,
\end{equation}
as \cite{polson2011data} showed that, $x \sim \gig(1/2, a, a/b^2)$ is equivalent to $x^{-1} \sim \ig(\lvert b \rvert, a)$.
Notably, RBHMC reverts to BHMC when $C \to 0^+$. 

\subsection{Complete Inference}
We review the sampling steps for the random variables in the BHMC and hence connect the sampling steps for all variables and show a full algorithm.

One could assume the base measure $H$ to be a multivariate normal distribution such that $H = \N(\bm{\mu}_0, \bm{\Sigma}_0)$.
To apply posterior conjugacy, one could then also set $F$, which samples the observation along with $\theta$, to a normal distribution with known covariance matrix.
That is, $\theta_k \coloneqq \bm{\mu}_k$ where $\bm{\mu}_k$ is the normal mean for the $k$th component, and $f(\x_n; \bm{\mu}_k)$ is then the density function corresponding to the normal distribution.
We write $N_{z k}$ for the number of observations that pass through $z$ and are assigned to component $k$.
Furthermore, we write $N_k$ the number of observations that are globally assigned to $k$.
Finally,~\Cref{tb:sample_bhmc} summarizes the sampling detail in the BHMC.

\begin{table}[!h]
\begin{center}
\caption{Sampling detail of the BHMC}
\label{tb:sample_bhmc}
\begin{tabular}{| m{2.5cm}| m{3cm}| m{5cm}|}
\hline
\textbf{Description} & \textbf{Variable} & \textbf{Posterior sampling detail} \\
\hline
\hline
Path assignment & $\m{V} = \{\bm{v}_n\}_{n=1}^N$ & omit since it has been modified by the RBHMC \\
\hline
Component assignment & $\m{c} = \{c_n\}_{n=1}^N$ & $c_n \propto \begin{cases}
\beta_{z k} f(\x_n; \theta_k) & existing~k \\
\beta_{z}^* f^*(\x_n) & new~k
\end{cases}$ \\
\hline
Component weights \newline (root node) & $\bm{\beta}_{z_0}$ & $\bm{\beta}_{z_0} \sim \Dir(N_{z_0 1}, \dots, N_{z K}, \gamma_0)$ \\
\hline
Component weights \newline (non-root node) &
$\tilde{\m{B}} = \{ \bm{\beta}_{z'} \}_{z' \in \nodeset \setminus \{z_0\}}$ &
$z = \parent(z')$ \newline
$\hat{N}_{z' k} = N_{z' k} + \gamma \beta_{z k}$ \newline
$\bm{\beta}_{z' k} \sim \Dir(\hat{N}_{z' 1}, \dots, \hat{N}_{z' K}, \gamma \beta^*_{z})$ \\
\hline
Component parameters & $\bm{\theta} = \{ \bm{\mu}_k \}_{k=1}^K$ &
$\bar{\bm{\x}}_k = \frac 1 {N_k} \sum_{n: c_n = k} \x_n$ \newline
$\tilde{\bm{\mu}}_k = \tilde{\bm{\Sigma}}_k \lp \bm{\Sigma}_0^{-1} \bm{\mu}_0 + N_k \bm{\Sigma}^{-1} \bar{\x}_k \rp $ \newline
$\tilde{\bm{\Sigma}}_k = \lp \bm{\Sigma}_0^{-1} + N_k \bm{\Sigma}^{-1} \rp^{-1}$ \newline
$\bm{\mu}_k \sim \N(\tilde{\bm{\mu}}_k, \tilde{\bm{\Sigma}}_k)$ \\
\hline
\multicolumn{3}{l}{*Refer to~\cite{Huang2021inferring} for a full analysis} \\
\end{tabular}
\end{center}
\end{table}

\Cref{alg:regmh-sampling} depicts the entire MCMC procedure.
The first part of the algorithm focuses on the assignment variable sampling.
For each observation, it proposes a path and the corresponding auxiliary variables and applies the MH scheme to probabilistically accept the proposal.
After the path assignment is updated, the corresponding auxiliary variable $\lambda_n$ will be sampled.

It is possible to sample a new component beyond the current number $K$ when updating $c_n$ for a certain $n$.
In that case, every node repeats the stick-breaking process for one more step according to the HDP machinery~\cite{Teh2006hier}.
At the root node, we obtain $\beta_{z_0 (K+1)} = \beta' \beta_{z_0}^*$ where $\beta' \sim \Beta(1, \gamma_0)$ and $\beta_{z_0}^* = 1 - \sum_{k=1}^{K+1} \beta_{z_0 (K+1)}$.
For any other node with parent $z$, we sample the new proportion for breaking the remaining stick by $\Beta(\gamma \beta_{z (K+1)}, \gamma \beta_{z}^*)$ and handle the rest in the same manner as that for the root.
We then update $K$ by $K+1$, and sample $(\m{B}, \bm{\eta})$ which are associated with each node $z$.
Finally, $\bm{\theta}$ are updated with regard to the $K$ components.
\begin{algorithm}[!t]
\caption{\textsc{Posterior Sampling for the RBHMC}}
\label{alg:regmh-sampling}
\begin{algorithmic}[1]
\State Initialize a tree
\Repeat
\For{$n \in \textsc{Shuffle}(\{1, \ldots, N\})$}
\State Sample a path $\bm{v}'_n$ and the corresponding $\bm{\beta}$ if needed
\State $s_{n}' \gets \argmax_{1 \le \ell \le L, z \in \sibs(v_{n \ell})} \rho_{n \ell z}$
\State Accept the proposal with probability $\mathcal{A}$
\State Update ${\lambda}_{n}$
\EndFor
\State Update $c_n$ for $n = 1,\dots, N$
\State Update $\bm{\beta}_{z}$, $\bm{\eta}_z$ for $z \in Z$
\State Update $\theta_k$ for $k = 1, \dots, K$
\Until{converged}
\end{algorithmic}
\end{algorithm}

\subsection{Output hierarchy}
\cite{adams2010tree,Huang2021inferring} propose to output the hierarchies with the highest complete data log likelihood.
For estimation or application, one might prefer presenting the hierarchy with the maximal posterior or likelihood during a finite number of MCMC draws.
However, there are interesting discussions on replacing the maximal posterior with other well-established criteria for clustering, e.g.~\cite{rastelli2018optimal,wade2018bayesian}, etc.
Their research focus is on flat clustering and we leave as a future research challenge to explore a better Bayesian criterion for choosing the output hierarchy.

We follow~\cite{adams2010tree,Huang2021inferring} and still apply the complete data likelihood to select the output hierarchy.
This approach applies the regularized complete data log likelihood (RCDLL) which is defined as
\begin{align}
\label{eq:rcdl}
\mathrm{RCDLL} = \sum_n \log p(\x_n, {c}_n, \bm{v}_n \mid \model \setminus \{ \m{c}, \m{V} \}) - 2C \max(0, \rho_{n s_n}) \, .
\end{align}
For presentation, any node with no siblings will be merged with its parent, and the empty nodes containing no observations will be excluded.

\section{Experimental Study}
\label{sec:experiment}
This study uses the \texttt{Animals} dataset~\citep{kemp2008discovery} and the \texttt{MNIST-fashion} dataset~\citep{xiao2017fmnist}, as evaluated in~\citet{Huang2021inferring}.
These datasets allow for intuitive interpretation of hierarchies, enabling a straightforward comparison between RBHMC and BHMC.
\texttt{MNIST-fashion} is sub-sampled with a batch of only 100 items for better qualitative outcomes.
Principal Component Analysis~\citep{hastie2009elements} reduces the data dimension to 7 and 10 for the two datasets, respectively.\footnote{The code and processed data are available at \url{https://bitbucket.org/weipenghuang/regbhmc}, and the GIG implementation is from \url{https://github.com/LMescheder/GenInvGaussian.jl}.}
For each dataset, we will conduct sequential analyses, including convergence, sensitivity, case, and co-occurrence analysis.
The convergence analysis will focus on RCDLL.
Sensitivity analysis aims to understand the impact of hyperparameters on performance.
We will discuss its experimental design in the next subsection. In the case analysis, we will compare hierarchies qualitatively, illustrate and quantitatively show clustering-focused F-measure performance.
This F-measure~\citep{steinbach2000comparison} is a modified version of the metric used to evaluate clustering quality, with implementation provided in~\Cref{sec:f-measure}.
Finally, co-occurrence analysis will examine clustering uncertainty at each  layer.

\bmhead{Design of sensitivity analysis}
This analysis focuses on the regularization hyperparameters $C$ and $\varepsilon_0$, while taking into account a hyperprior for the PR imposed variable $\bm{\eta}$.
An overly large value of $C$ would dominate the regularization term and result in uniform assignment, whereas a very small value would make the model similar to the original BHMC.
In interpretation, $\varepsilon_0$ can be interpreted as the largest unnormalized margin between the closest observation and the hyperplane $\bm{\eta}$.
A small value of $\varepsilon_0$ may not constrain the model effectively, and a large value may penalize certain path assignments unnecessarily.
The analysis aims to explore the sensitivity of the hyperparameters and the approach's capability to obtain improvements.

Hence, the analysis aims to explore ($i$) the sensitivity of the hyperparameters; ($ii$) the approach's capability to obtain improvements.
We examine two simple measures, which are
\begin{itemize}
\item
\textbf{Average Inner Distance} (AID):
the average empirical squared $L_2$ distance within the nodes,

\item
\textbf{Average Outer Distance} (AOD):
the average centroid squared $L_2$ distance between the siblings.
\end{itemize}
These scores will be only averaged over the non-empty nodes.
Regarding each hyperparameter configuration, the analysis is repeated $30$ times.
We then write $\mathrm{AID}$ as the expectation over $\mathrm{AID}_{single}$, and likewise for AOD.
Specifically,
\begin{align*}
  \mathrm{AID}_{single} &= \frac 1 {|\nodeset \setminus \{ z_0 \}|}\sum_{z \ne z_0 } \frac {2 \sum_{ \x_{n'}, \x_{n} \in z} \mathbbm{1}(n > n') \lVert \x_{n} - \x_{n'} \rVert^2_2 } {N_z (N_z -1)} \, .
\end{align*}
Let $\bar{\x}_{z}$ denote the centroid in a node $z$ where $\bar{\x}_{z} = \frac{1}{N_{z}} \sum_{\x_n \in z} \x_n $.
We write $\mathcal{SP}$ to denote a set maintaining the ordered \emph{sibling pairs} under any parent node in the tree.
AOD is defined by
\begin{align*}
  \mathrm{AOD}_{single} &= \frac{1} {|\mathcal{SP}|} \sum_{(z, z') \in \mathcal{SP}} \lVert \bar{\x}_{z} - \bar{\x}_{z'} \rVert_2^2  \,.
\end{align*}


It is possible to generate a random tree with only one child under each parent, known as a singular path, which can cause AOD to fail.
However, any sampling steps that result in this scenario after the burn-in phase will be disregarded.
Due to the \emph{the-rich-get-richer} property of nCRP, once a singular sample is drawn, subsequent samples in the chain are more likely to be singular.
Fortunately, these occurrences are extremely rare.

\subsection{Animals data}
First of all, we fix the hyperparameters to $\alpha=0.35, \gamma=1, \gamma_0=0.85, L=3,
\nu_0=1, F=\N(\cdot, \m{I}), H=\N(\bm{0}, \I)$ which roughly follows the settings in~\citet{Huang2021inferring}.

\begin{figure}[h]
\centering
\includegraphics[width=0.82\textwidth]{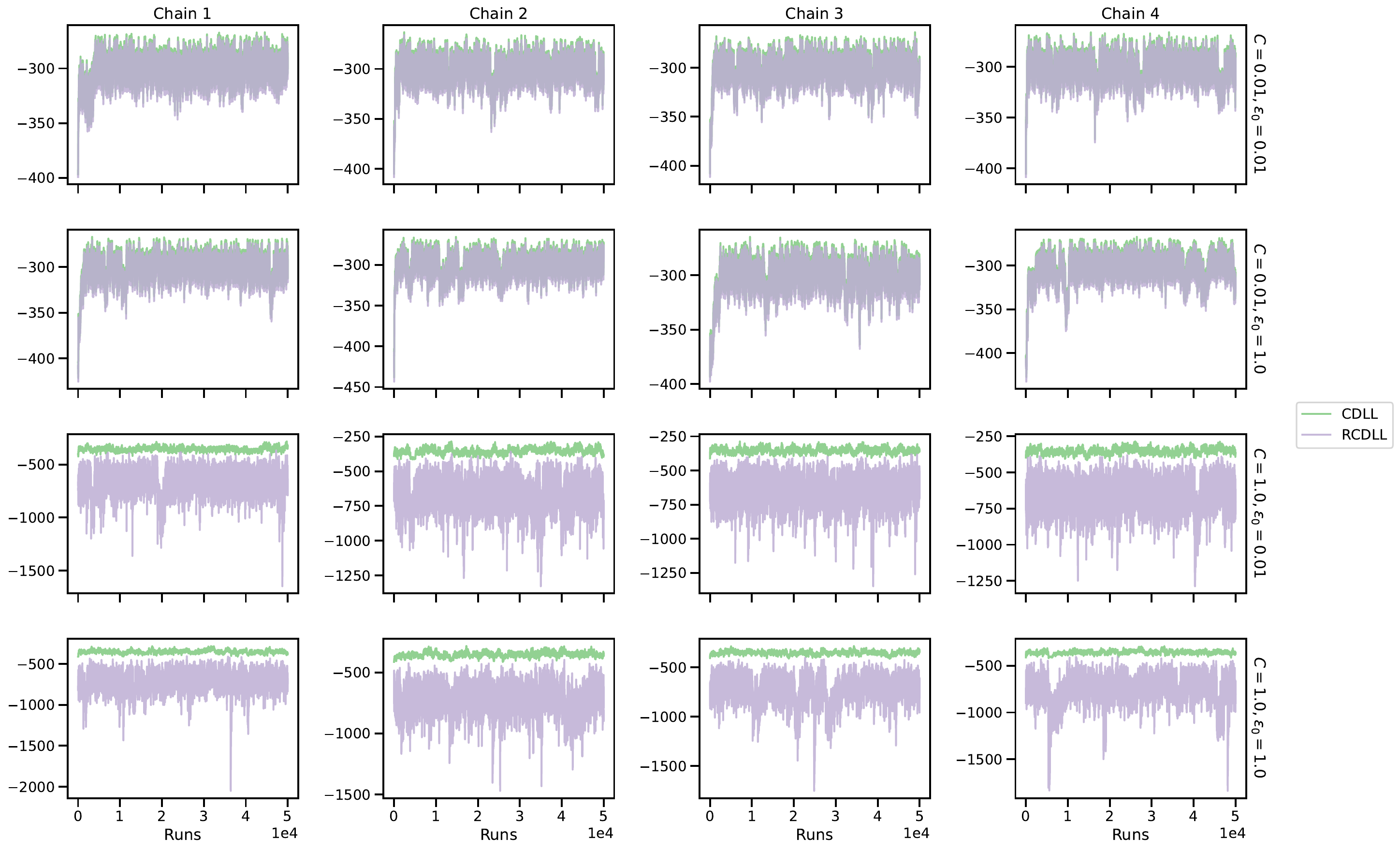}
\caption{Convergence analysis on \texttt{Animals}} \label{fg:conv_animals}
\end{figure}

\begin{figure}[h]
\centering
\includegraphics[width=0.75\textwidth]{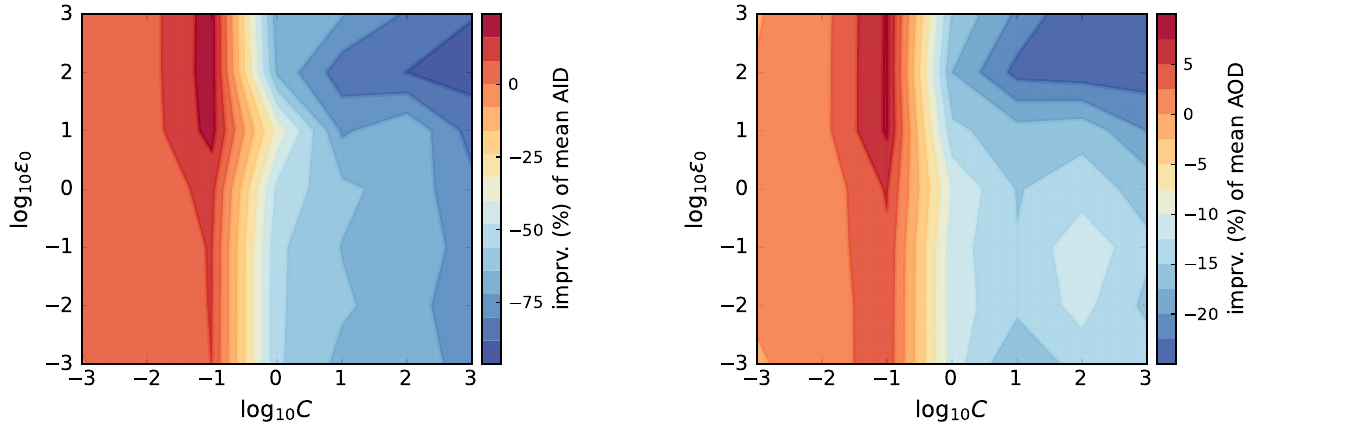}
\caption{Sensitivity analysis on \texttt{Animals}. The percentage of improvement (imprv. \%) is computed against the results from the BHMC for each measure} \label{fg:sens_animals}
\end{figure}

\begin{figure}[h]
\centering
\subfloat[Output of the BHMC from chain 1]{
\includegraphics[width=0.35\textwidth]{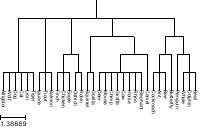}
}
\quad
\subfloat[Output of the BHMC from chain 2]{
\includegraphics[width=0.35\textwidth]{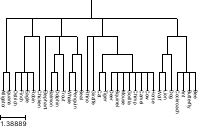}
}
\\
\subfloat[Output of the RBHMC with $C=0.1, \varepsilon_0=0.1$ from chain 1]{
\includegraphics[width=0.35\textwidth]{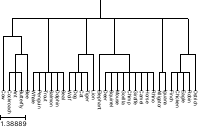}
}
\quad
\subfloat[Output of the RBHMC with $C=0.1, \varepsilon_0=0.1$ from chain 2]{
\includegraphics[width=0.35\textwidth]{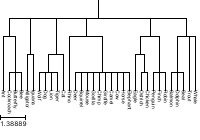}
}
\\
\subfloat[Output of the RBHMC with $C=0.1, \varepsilon_0=10.0$ from chain 1]{
\includegraphics[width=0.35\textwidth]{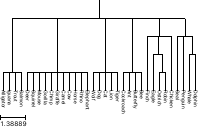}
}
\quad
\subfloat[Output of the RBHMC with $C=0.1, \varepsilon_0=10.0$ from chain 2]{
\includegraphics[width=0.35\textwidth]{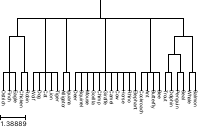}
}
\caption{Examples of output trees on \texttt{Animals} with different settings. The trees are selected by the highest CDLL and RCDLL accordingly. The rows are for each MCMC chain and the column will specify a certain level.}
\label{fg:animal_tree_examples}
\end{figure}

\subsubsection{Convergence analysis of the log likelihoods}
\Cref{fg:conv_animals} shows the complete data log likelihood (CDLL) and the RCDLL, i.e. ~\Cref{eq:rcdl}.
All plots display that both statistics increase to a certain level while oscillating as more runs are carried out.
The variable $C$ is one key factor that influences the range of change for the RCDLL.
When $C=0.01$, the RCDLL and CDLL are rather similar; however, for $C=1$, the fluctuations of the RCDLL become a lot stronger.
A number of plots demonstrate that the RCDLL still fluctuates more strongly than the CDLL.
This implies that the regularization searches the space more effectively but the CDLL is maintained at a rather consistent level.

\subsubsection{Sensitivity analysis}

\Cref{fg:sens_animals} illustrates the sensitivity analysis for two hyperparameters.
Improvement is defined as the percentage increase in performance over BHMC for AID and AOD. We collected samples every fifth draw from 50,000 draws, with 25,000 burn-in runs.
Performance of each measure had a non-linear correlation with hyperparameters.
RBHMC outperformed BHMC with various hyperparameter settings, e.g., $C=0.1$ and $\varepsilon_0=1$ showed growth on AID and AOD.
Improvements were mainly seen when $C \le 0.1$ or $\log_{10} C \le -1$.
While $C$ was more influential, a suitable $\varepsilon_0$ was still important for further improvement.

The percentage improvement is computed with the BHMC result as the baseline.
It should be noted that for AID, the lower the value, the better, while for AOD, the higher the value, the better.
Numerically, we observe that the improvement of mean AID can reach around $20\%$-$25\%$ and the mean AOD can improve to about $10\%$.

\begin{figure}[h]
\centering
\includegraphics[width=0.5\textwidth]{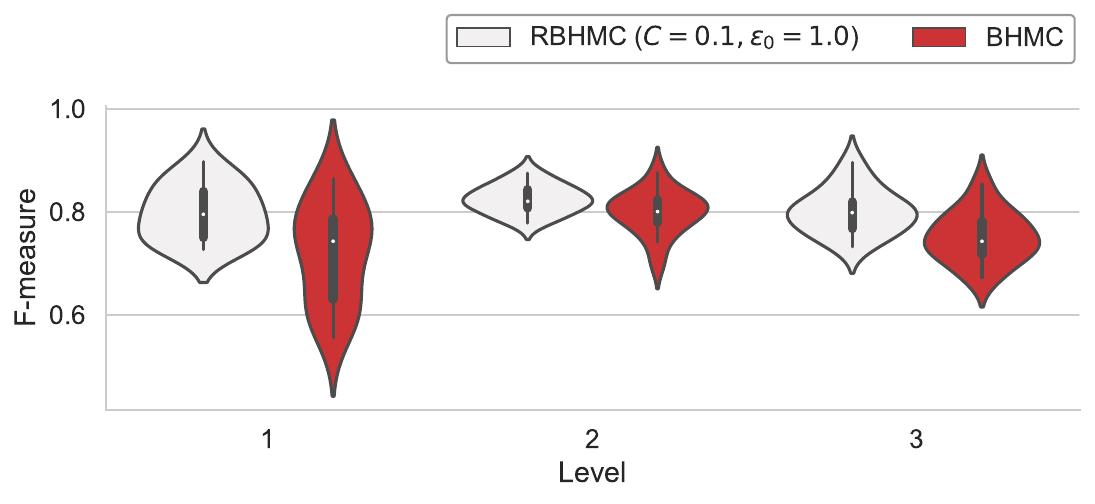}
\caption{F-measure comparison for \texttt{Animals} on two configurations, for which 15 parallel chains are run with $25,000$ burn-in runs and $50,000$ samples for selecting the highest CDLL and RCDLL}
\label{fg:animals_fmeasure}
\end{figure}

\begin{figure}[h]
\centering
\includegraphics[width=0.85\textwidth]{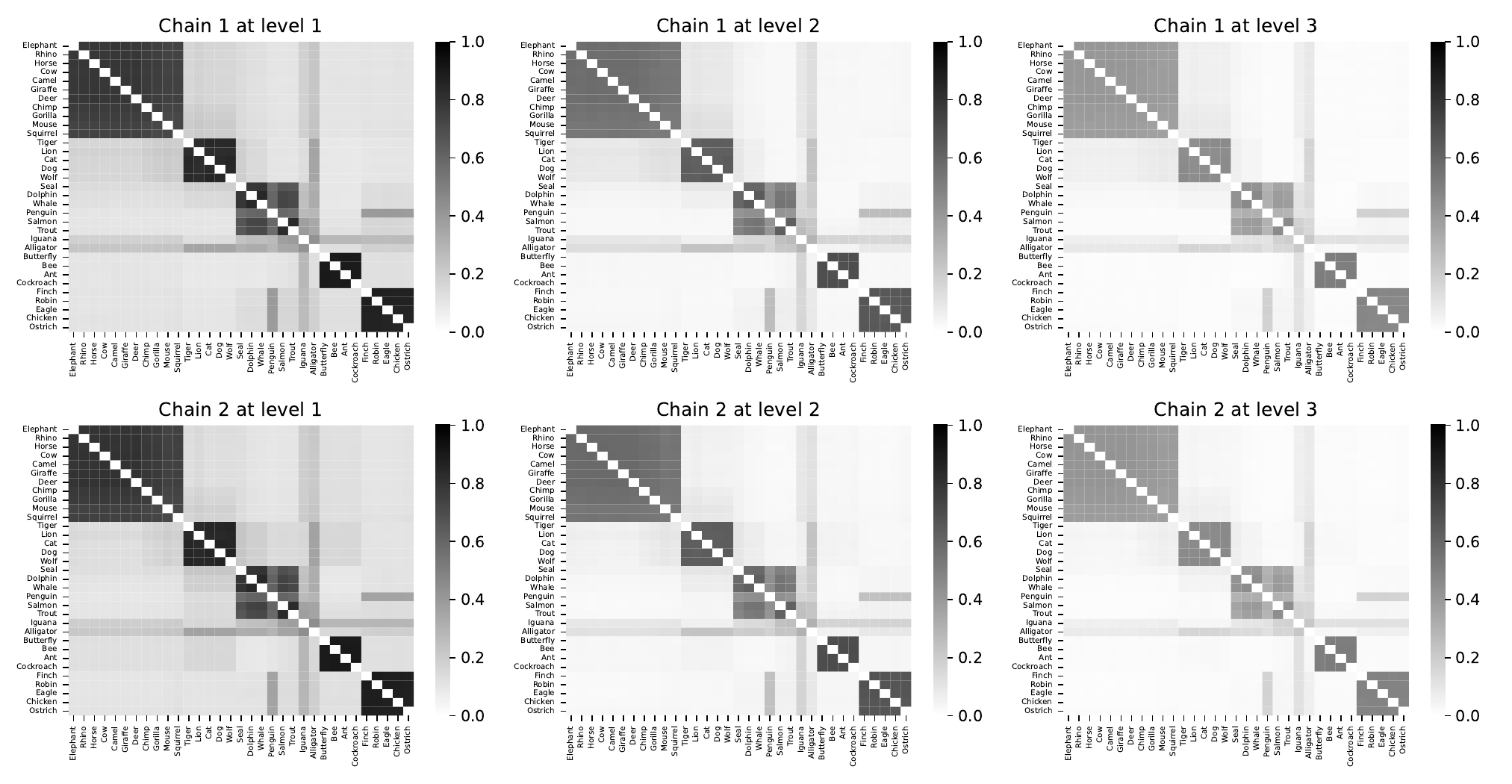}
\caption{Co-occurrence plots for the BHMC on \texttt{Animals} for two independent chains. The rows are for each MCMC chain and the column will specify a certain level.}
\label{fg:cooc_animals_bhmc}
\end{figure}

\begin{figure}[h]
\centering
\includegraphics[width=0.85\textwidth]{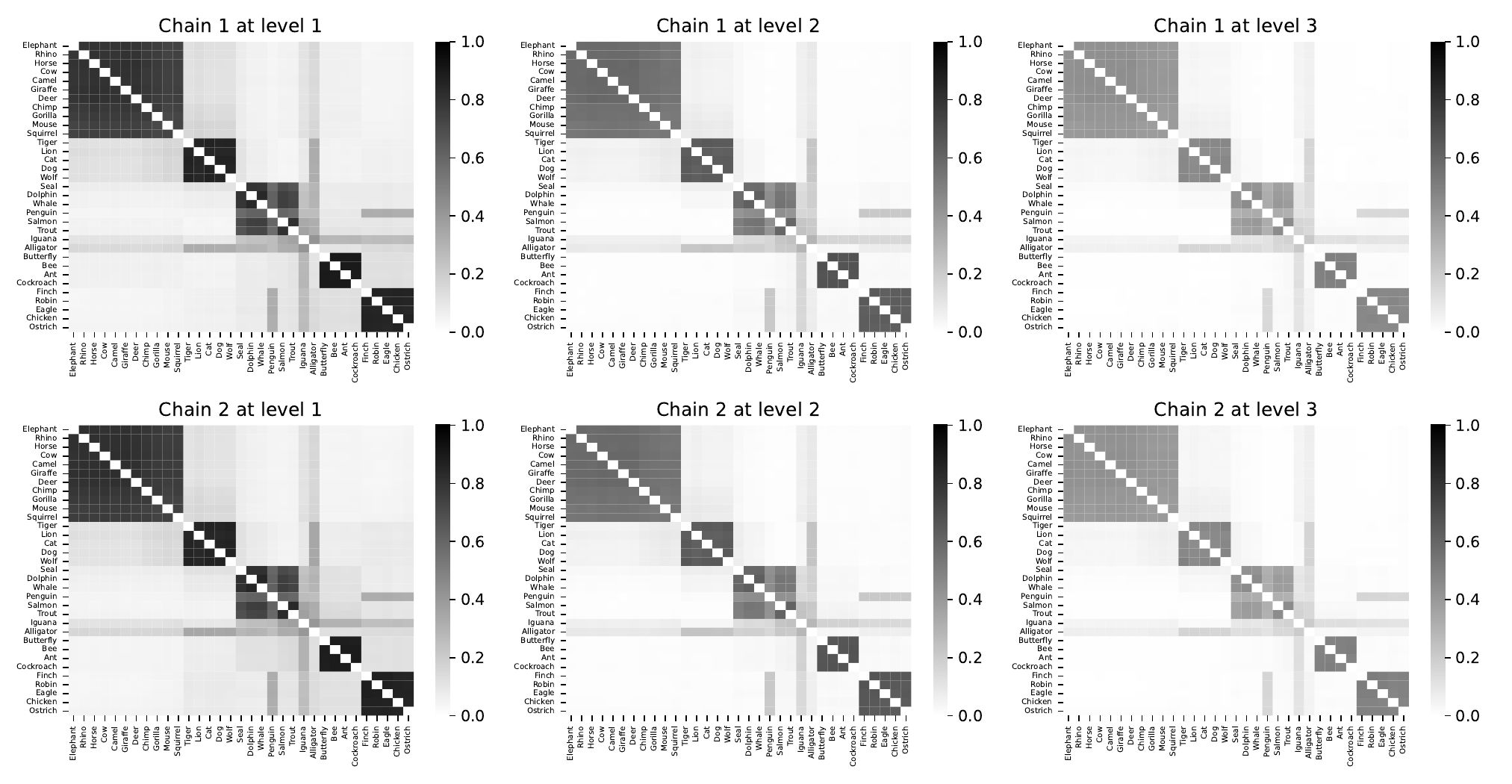}
\caption{Co-occurrence plots for the RBHMC on \texttt{Animals} for two independent chains, with $C=0.1$ and $\varepsilon_0=0.1$. The rows are for each MCMC chain and the column will specify a certain level.}
\label{fg:cooc_animals_rbhmc1}
\end{figure}

\begin{figure}[h]
\centering
\includegraphics[width=0.85\textwidth]{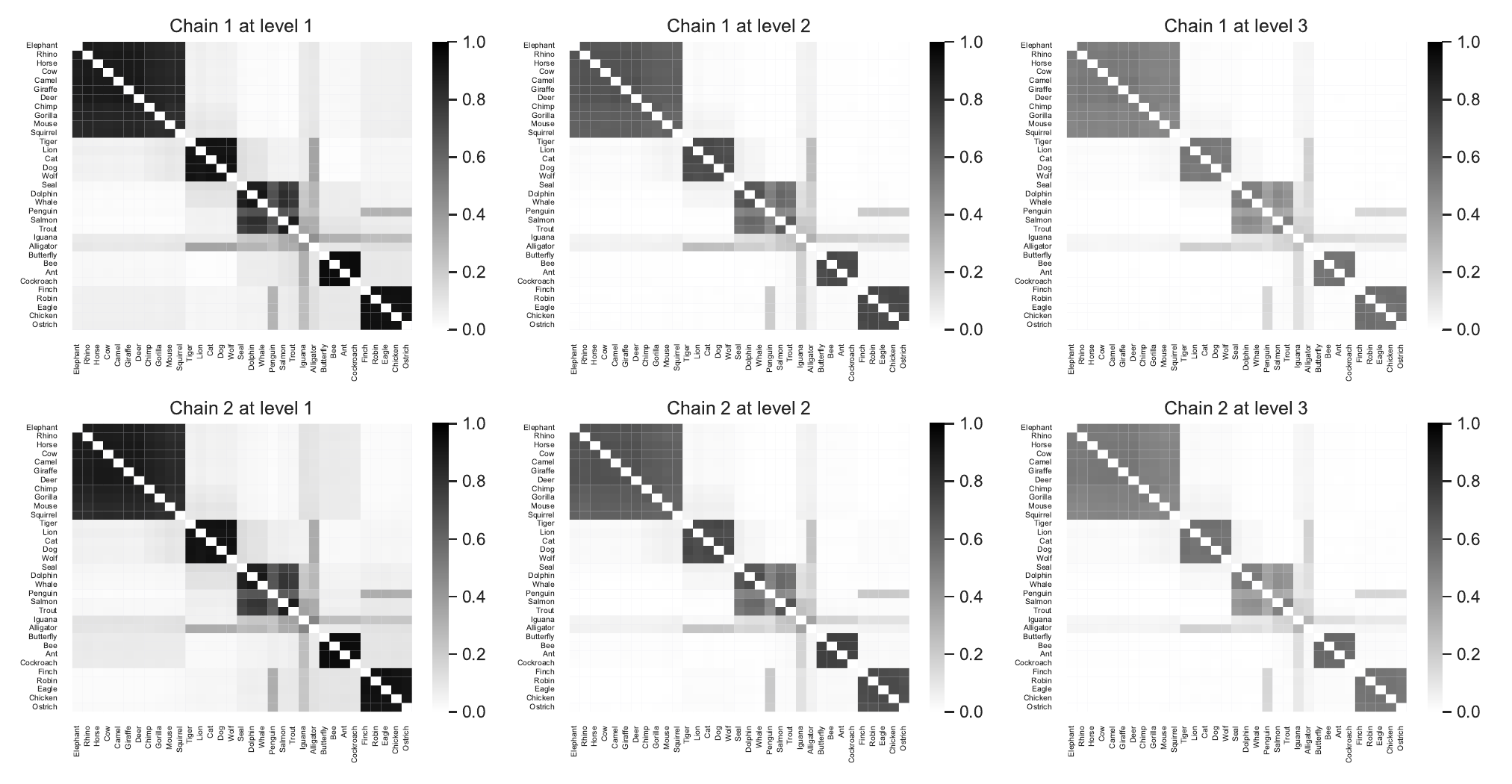}
\caption{Co-occurrence plots for the RBHMC on \texttt{Animals} for two independent chains, with $C=0.1$ and $\varepsilon_0=10.0$. The rows are for each MCMC chain and the column will specify a certain level.}
\label{fg:cooc_animals_rbhmc2}
\end{figure}

\subsubsection{Case analysis}
In this section, we test several hyperparameter settings with two separate chains.
At each chain, we run the first $25,000$ draws as burn-in, and report the hierarchy with the highest RCDLL in the following $50,000$ draws.
For the BHMC, the hierarchy with the highest CDLL is reported.


In~\Cref{fg:sens_animals}, we learned that $(C=0.1, \varepsilon_0=0.1)$ and $(C=0.1, \varepsilon_0=10)$ are relatively good hyperparameter value pairs.
In~\Cref{fg:animal_tree_examples}, we present two hierarchies generated under these configurations and two hierarchies from the BHMC model.
BHMC performed well overall, but there are still some flaws, such as the first example of the BHMC result.
However, a small number of misplaced items are understandable and should be allowed in a probabilistic clustering.
The hierarchies of the RBHMC with $(C=0.1, \varepsilon_0=0.1)$ form more reasonable structures.
For $(C=0.1, \varepsilon_0=10)$, the trees tend to approach a flat clustering result, which may be due to the strong $\varepsilon_0$ playing a role in constraining the solutions.
Although $(C=0.1, \varepsilon_0=10)$ is a well-performing pair of hyperparameters when only considering AOD and AID in~\Cref{fg:sens_animals}, the hierarchical ``shape'' should also be considered in selecting good hyperparameters.
From this perspective, the setting with a smaller $\varepsilon_0$ may be preferable.


Despite the BHMC's strong performance at lower levels, it may produce a disordered cluster combination at higher levels due to the nature of HDPMM.
We investigate whether the RBHMC can outperform BHMC in terms of F-measure, especially for nodes at higher levels.
To handle the unlabeled \texttt{Animals} dataset, we manually categorize it into eight classes: \emph{birds}, \emph{land mammals}, \emph{predators}, \emph{insects}, \emph{amphibians}, \emph{water animals}, \emph{mice}, and \emph{fish} (the complete labeling is available in~\Cref{sec:animal_label} for verification).
We then compare the F-measure of RBHMC and BHMC against the clustering at each level.
RBHMC with $(C=0.1, \varepsilon_0=1)$ and BHMC are run with 10 chains each.

%

\Cref{fg:animals_fmeasure} compares the F-measure at different levels for two algorithms.
The results indicate that at the first level, the RBHMC algorithm performs better overall, although there are cases where BHMC performs better.
At lower levels, the RBHMC outperforms BHMC, contrary to the original expectation of equal performance.
This suggests that better clustering at higher levels can lead to better quality clustering at lower levels.

\subsubsection{Co-occurrence analysis}

We present co-occurrence plots of hierarchies sampled using MCMC through different levels.
The BHMC results in~\Cref{fg:cooc_animals_bhmc} clearly show co-occurrence uncertainty in higher levels, with the heatmap for the first level filled with gray, indicating a co-occurrence probability of about 0.35.
On the other hand, regularization significantly reduces uncertainty in higher levels, as demonstrated in~\Cref{fg:cooc_animals_rbhmc1} and~\Cref{fg:cooc_animals_rbhmc2}.
This confirms that the PR approach effectively reduces nodal variance, particularly for higher nodes in the tree.
%

The RBHMC with $(C=0.1, \varepsilon_0=0.1)$ (\Cref{fg:cooc_animals_rbhmc1}) has slightly higher uncertainty at higher levels compared to the approach using $(C=0.1, \varepsilon_0=10)$ (\Cref{fg:cooc_animals_rbhmc2}).
In~\Cref{fg:cooc_animals_rbhmc2}, the dark areas are deeper than those in~\Cref{fg:cooc_animals_rbhmc1}, and there is less light gray outside these clusters.
The qualitative analysis suggests that $(C=0.1, \varepsilon_0=10)$ performs better by potentially creating more single nodes under one parent.

As shown above, both strong $C$ and $\varepsilon_0$ are inclined to reduce the ``structure'' of the hierarchy.
This means that if $C$ or $\varepsilon_0$ are excessively strong, it may remove too much variability and transform the tree into a flat cluster or even a single path.
Practitioners might consider this trade-off when constructing hierarchies in future applications.

\subsection{MNIST-fashion data}
For this dataset, part of the hyperparameters are set as $\alpha=0.2, \gamma=1.5, \gamma_0=0.85, L=4, \nu_0=1, F=\N(\cdot, \m{I}), H=\N(\bm{0}, \I)$.

\subsubsection{Convergence analysis of the log likelihoods}
\begin{figure}[!ht]
\centering
\includegraphics[width=0.82\textwidth]{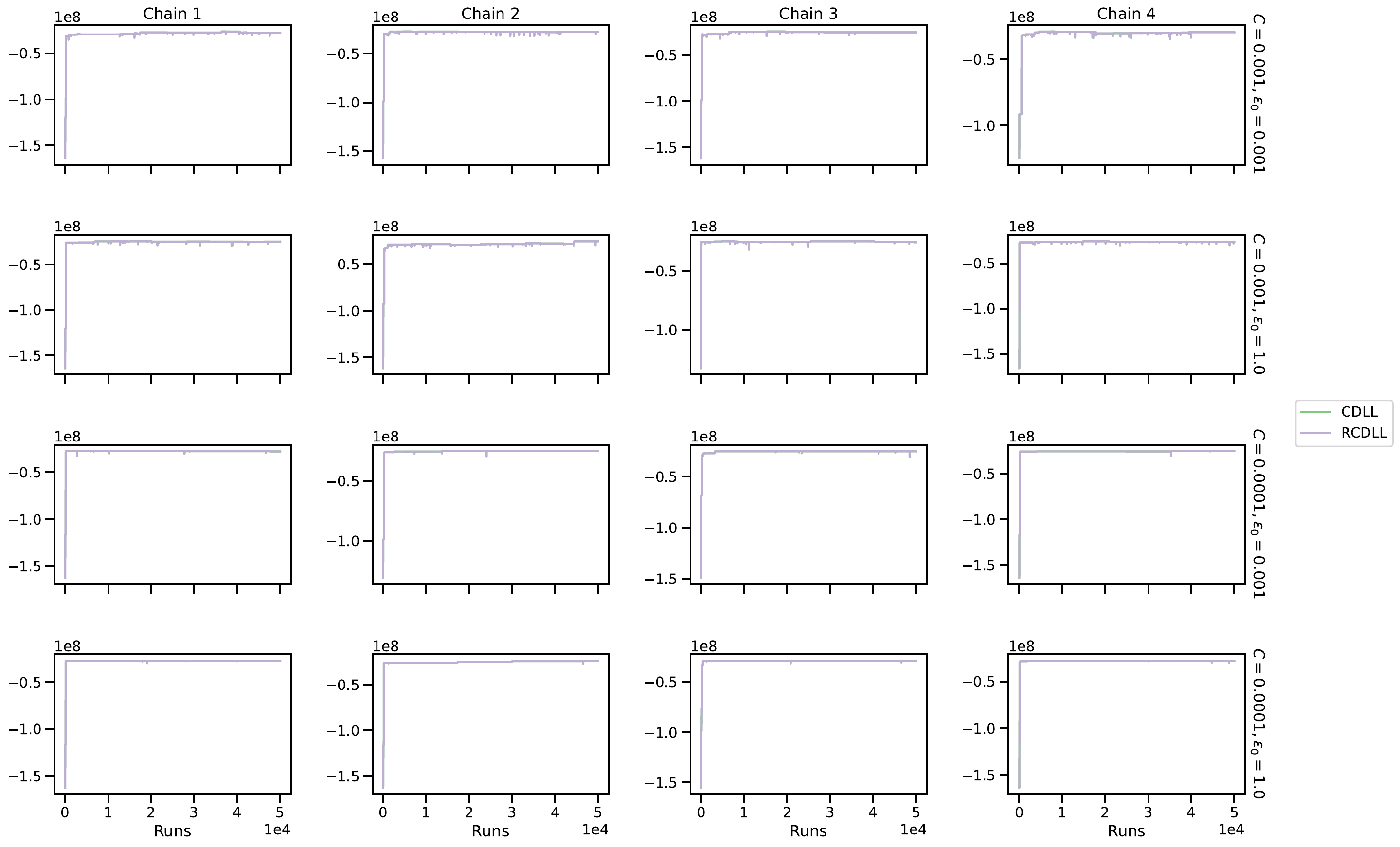}
\caption{Convergence analysis on \texttt{MNIST-fashion}} \label{fg:conv_fmnist}
\end{figure}

The CDLL and RCDLL of the algorithms for \texttt{MNIST-fashion} are demonstrated in~\Cref{fg:conv_fmnist}.
The four chains are run with $50,000$ draws.
In this example, we observe that both CDLL and RCDLL grow rapidly and keep fluctuating in all chains.
We are satisfied to see such fluctuations of the log-likelihoods as it means that the inference keeps exploring the solution space.

\begin{figure}[!ht]
\centering
\includegraphics[width=0.75\textwidth]{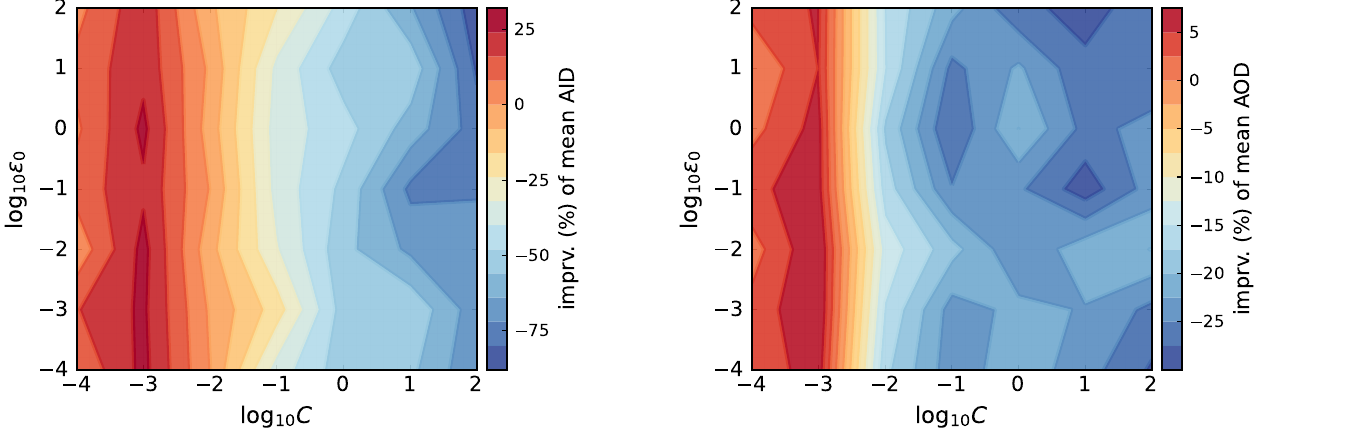}
\caption{Sensitivity analysis on \texttt{MNIST-fashion}. The percentage of improvement (imprv. \%) is computed against the results from the BHMC for each measure} \label{fg:sens_fmnist}
\end{figure}

\begin{figure}[!ht]
\centering
\subfloat[Output of the BHMC from chain 1]{
\includegraphics[width=0.85\textwidth]{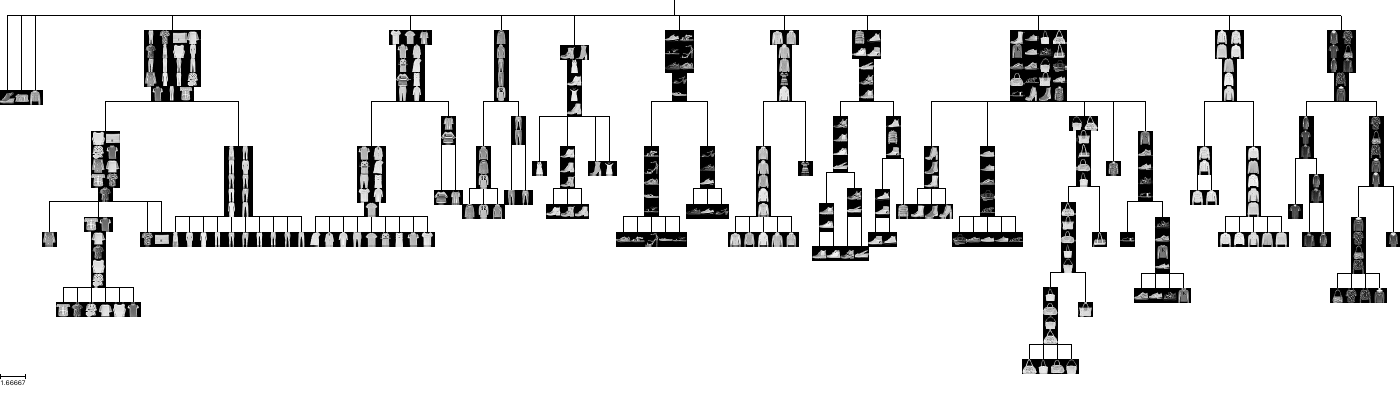}
}
\\
\subfloat[Output from RBHMC with $C=0.001, \varepsilon_0=0.01$ from chain 1]{
\includegraphics[width=0.85\textwidth]{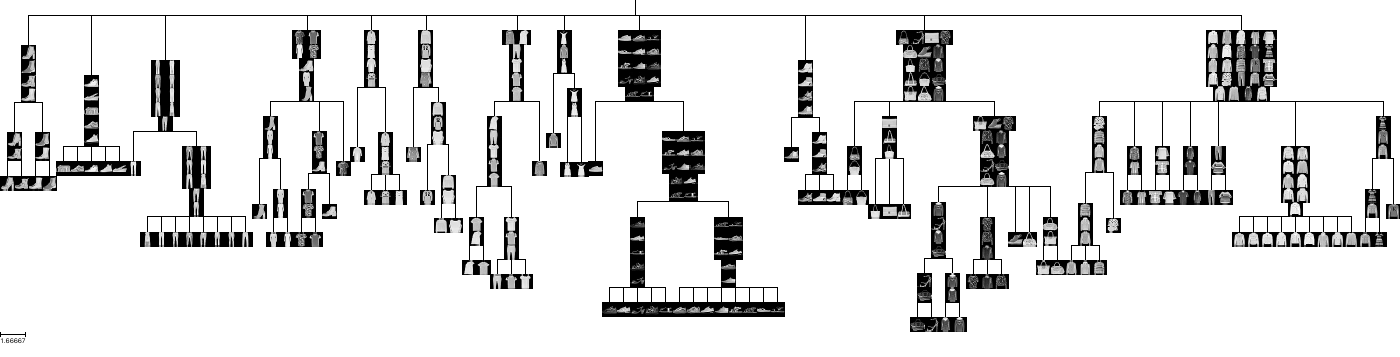}
}
\\
\subfloat[Output from the RBHMC with $C=0.001, \varepsilon_0=1$, from chain 1]{
\includegraphics[width=0.85\textwidth]{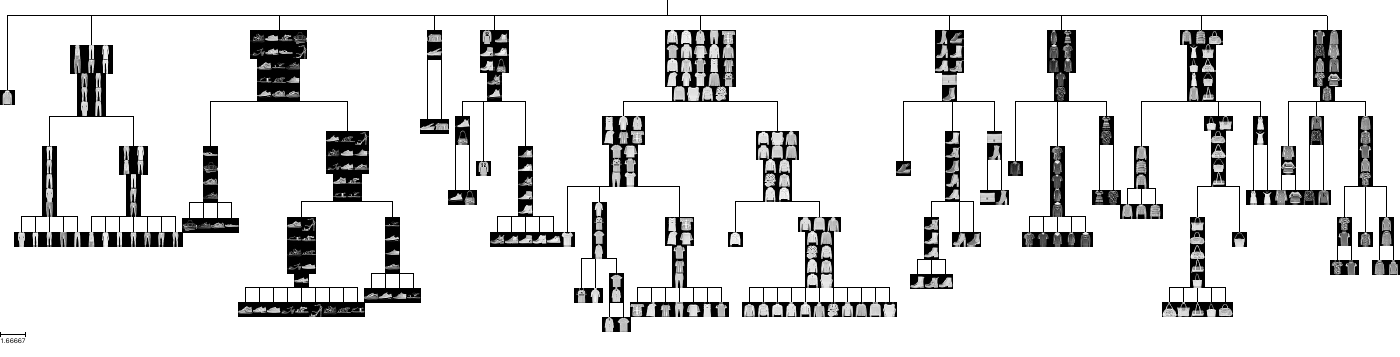}
\label{fg:fmnist_example_3}
}
\\
\subfloat[Output from the RBHMC with $C=1, \varepsilon_0=1$, from chain 1\label{fg:poor_example}]{
\includegraphics[width=0.85\textwidth]{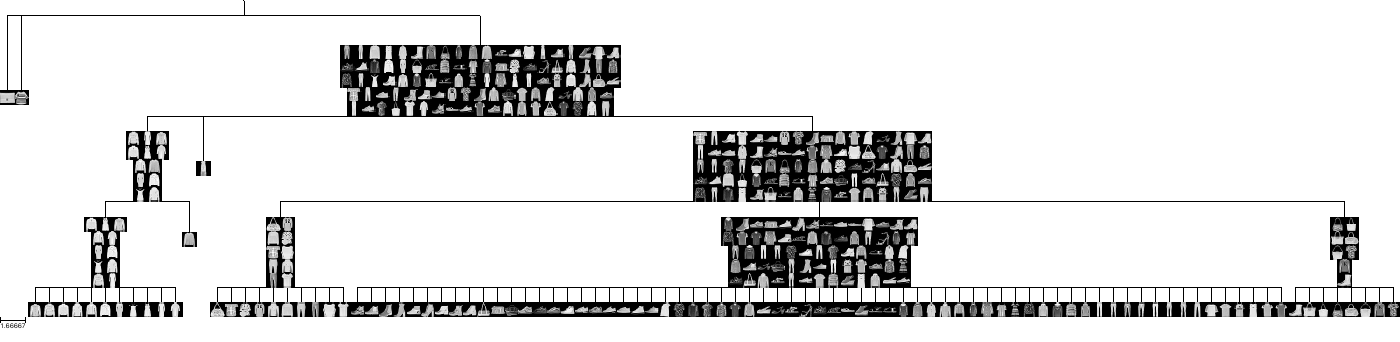}
}
\caption{Examples of output trees on \texttt{MNIST-fashion} with different settings}
\label{fg:fmnist_tree_examples}
\end{figure}

\begin{figure}[!ht]
\centering
\includegraphics[width=0.5\textwidth]{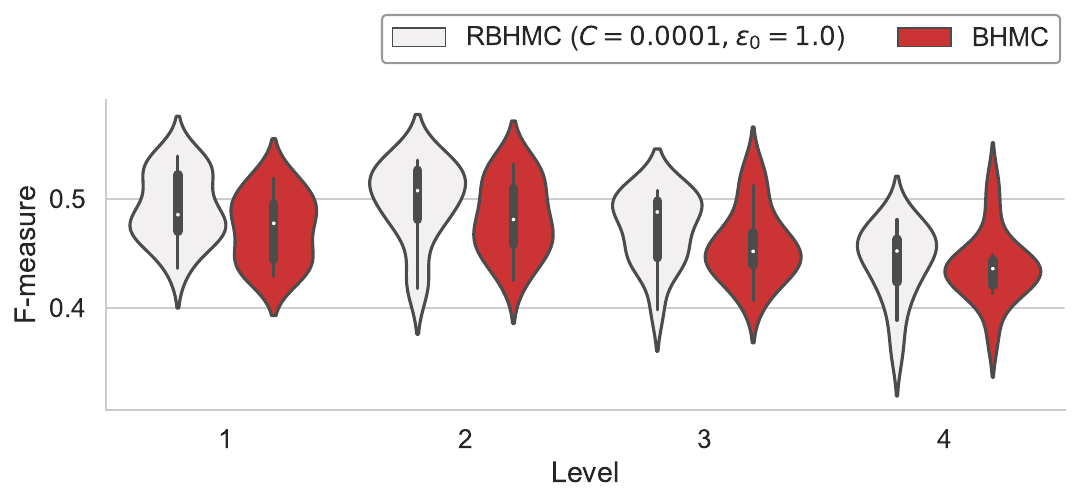}
\caption{F-measure comparison for \texttt{MNIST-fashion} on two configurations, for which 15 parallel chains are run with $25,000$ burn-in runs and $50,000$ samples for selecting the highest CDLL and RCDLL}
\label{fg:fmnist_fmeasure}
\end{figure}

\begin{figure}
\centering
\includegraphics[width=0.82\textwidth]{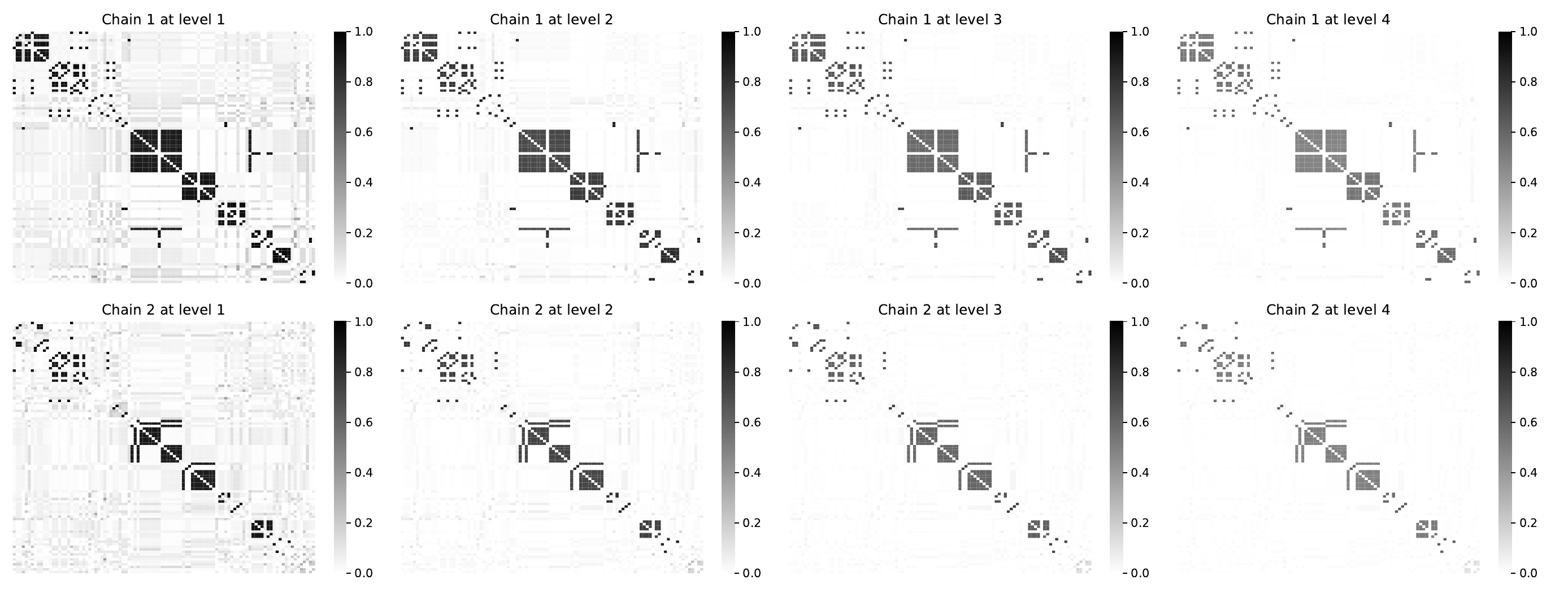}
\caption{Co-occurrence plots for the BHMC on \texttt{MNIST-fashion} for two independent chains. The rows are for each MCMC chain and the column will specify a certain level.}
\label{fg:cooc_fmnist_bhmc}
\end{figure}

\begin{figure}[!ht]
\centering
\includegraphics[width=0.82\textwidth]{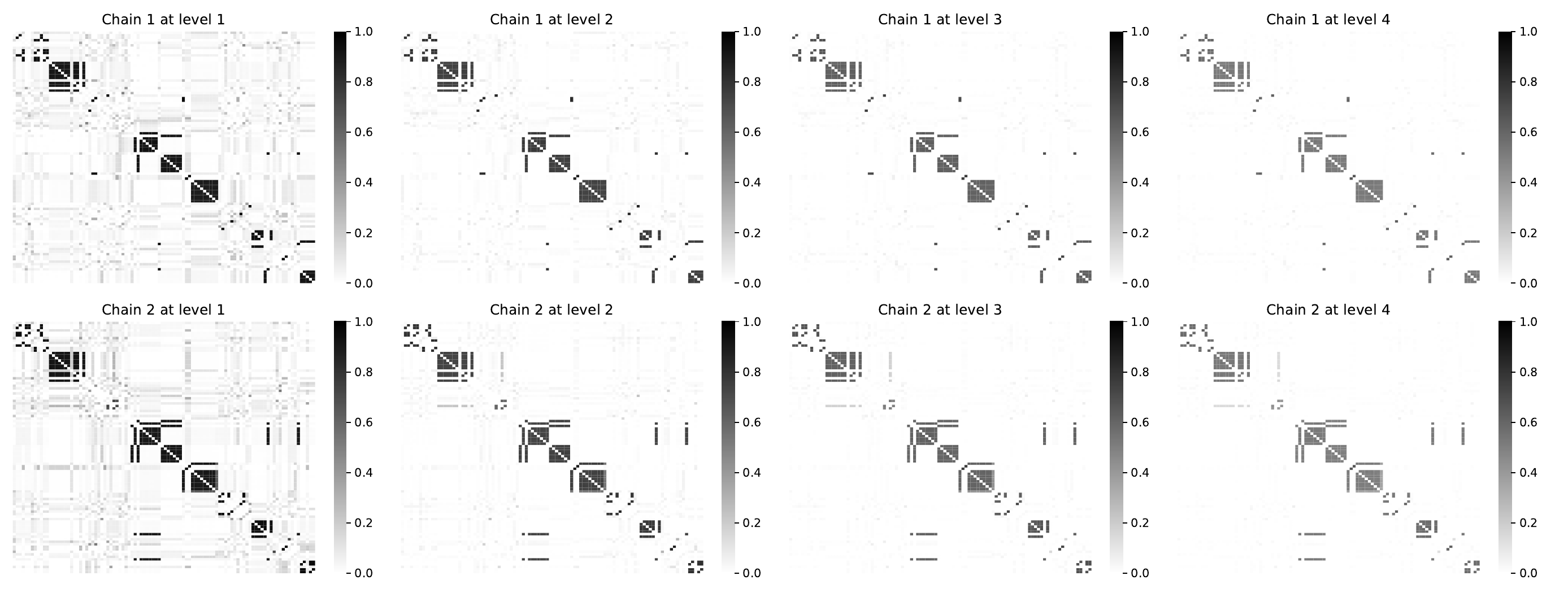}
\caption{Co-occurrence plots for the RBHMC on \texttt{MNIST-fashion} for two independent chains, with $C=0.001$ and $\varepsilon_0=1.0$. The rows are for each MCMC chain and the column will specify a certain level.}
\label{fg:cooc_fmnist_rbhmc1}
\end{figure}

\begin{figure}[!ht]
\centering
\includegraphics[width=0.82\textwidth]{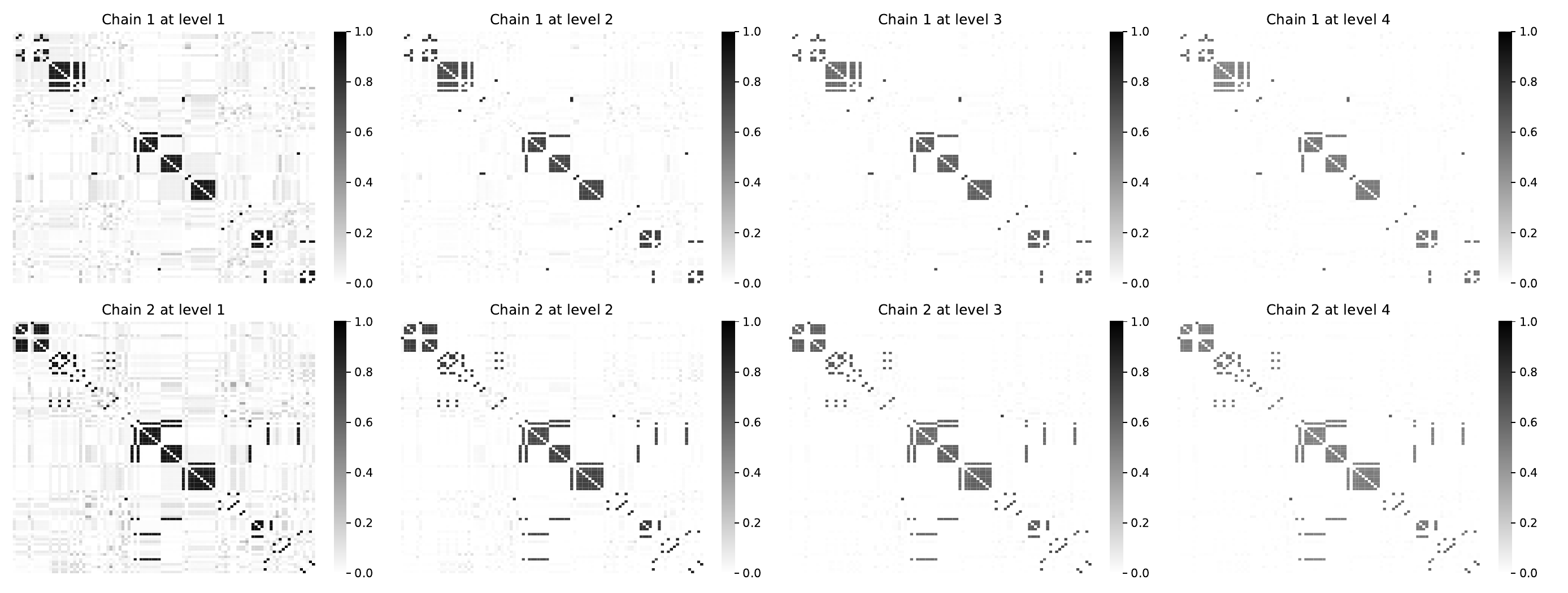}
\caption{Co-occurrence plots for the RBHMC on \texttt{MNIST-fashion} for two independent chains, with $C=0.001$ and $\varepsilon_0=0.01$. The rows are for each MCMC chain and the column will specify a certain level.}
\label{fg:cooc_fmnist_rbhmc2}
\end{figure}

\subsubsection{Sensitivity analysis}
The experiments are run for $50,000$ draws in this analysis and the results are shown in~\Cref{fg:sens_fmnist}.
Each $(C, \varepsilon_0)$ pair is repeated $30$ times but the rare cases of having a singular path are eliminated.
The scale of hyperparameters for achieving improvements are biased towards smaller values compared with \texttt{Animals}.
In agreement with the results for \texttt{Animals}, the improvement changes are smooth given the range of values.
Moreover, it again illustrates that the RBHMC can obtain notable improvements with the properly selected hyperparameters.
The change of $C$ affects the improvements more sensitively, though, choosing a suitable value for $\varepsilon_0$ is still crucial.
In this case, the mean AID decreases by around $27$-$28\%$ over the BHMC.
Moreover, the mean AOD performs about $7.5\%$ better than that of the BHMC.

\subsubsection{Case analysis}
We run the hyperparameter settings for two chains each of which will burn the first $25,000$ runs and draw from the following $75,000$ runs.
The generated trees are selected by the maximal RCDLL and are demonstrated in~\Cref{fg:fmnist_tree_examples}.

In~\Cref{fg:poor_example}, we provide a poorly performing example generated from the setting $C=1$ and $\varepsilon_0=1$, where overly restrictive constraints confine the solution to an undesirable region of the state space.
This scenario leads to irrational grouping of data due to the excessive constraints.
Ideally, the This scenario leads to irrational grouping of data due to the excessive constraints.
Ideally, RBHMC should ensure uniform cluster allocations with very large $C$. However, nCRP proposes clustering based on the \emph{the-rich-get-richer} principle, resulting in a biased hierarchy, unless it runs for an unrealistic number of iterations in practice.
The RBHMC should ensure uniform cluster allocations with very large $C$.
However, nCRP proposes clustering based on the \emph{the-rich-get-richer} principle, resulting in a biased hierarchy, unless it runs for an unrealistic number of iterations in practice.

Overall, the BHMC returns a reasonable hierarchy.
With regard to the RBHMC, we deliberately exhibit the results with $C=0.001$ from~\Cref{fg:sens_fmnist}, which is a value that leads to improvements.
The RBHMC results generally look better clustered, in particular the one with the hyperparameter setting $C=0.0001, \varepsilon_0=1$, in~\Cref{fg:fmnist_example_3}.

The \texttt{MNIST-fashion} data has ground-truth labels in categories that allows us to compute the F-measure directly.
Here, $15$ chains with $25,000$ burn-in rounds and $50,000$ draws are simulated.
\Cref{fg:fmnist_fmeasure} demonstrates that, at a higher level, the RBHMC has the potential to perform much better than the BHMC.
The distributions of the RBHMC's F-measure are also skewed towards a better direction.
At the lower levels, the RBHMC outperforms BHMC by an even greater amount.

\subsubsection{Co-occurrence analysis}
We apply the co-occurrence analysis on the MCMC samples, i.e., $50,000$ samples after the burn-in phase.
\Cref{fg:cooc_fmnist_bhmc} shows the result for two BHMC chains.
As expected, there are a lot more gray areas in the higher levels than that of~\Cref{fg:cooc_fmnist_rbhmc1,fg:cooc_fmnist_rbhmc2}.
In the RBHMC, the structures are more certain by realizing that there are dark blocks while the other areas are close to being completely white (i.e., $0\%$).
This again indicates that the regularization manages to constrain the solutions into the desired space.

\section{Conclusion}
\label{sec:conclusion}

To address the shortcoming of the BHMC model, which may have high nodal variance at higher levels of hierarchy, this paper proposes applying PR to impose max-margin constraints on the hierarchical structure.
The paper details the modeling and inference procedures for an improved Bayesian HC algorithm, called RBHMC. Empirical studies show its advantages over the original BHMC model and its ability to achieve desired improvements.
The authors suggest further investigation and application of this methodology to a wider range of Bayesian tasks.
Additionally, exploring other constraint space constructions and regularized functions would be beneficial.
Finally, extending the approach to handle large-scale problems is essential.

%





\bibliography{regbhmc}


\begin{appendices}
\section{Derivation for the Acceptance Probability}
\label{sec:rbhmc-ap-accept}

For computing the acceptance probability in the MH scheme, we obtain
\begin{align*}
&\frac {q(\model_0', \bm{\eta}, \bm{\lambda})q(\bm{v}_n, \m{B}, {s}_n \mid \bm{v}_n', \m{B}', {s}_n')} {q(\model_0, \bm{\eta}, \bm{\lambda})q(\bm{v}_n', \m{B}', {s}_n' \mid \bm{v}_n, \m{B}, {s}_n)} \\
&= \frac {\tilde{f} \left( \x_n, \lambda_{n}; \bm{v}_n', s_n', \m{B}', \bm{\theta}, \bm{\eta} \right) p(\m{V}') p(\m{B}') } {\tilde{f} \left( \x_n, \lambda_n;\bm{v}_n, s_n, \m{B}, \bm{\theta}, \bm{\eta} \right) p(\m{V}) p(\m{B}) }
\frac {q(\bm{v}_n \mid \bm{v}_n') q(\m{B}) } {q(\bm{v}_n' \mid \bm{v}_n) q(\m{B}') }  \\
&= \frac {\tilde{f} \left( \x_n, \lambda_{n}; \bm{v}_n', s_n', \m{B}', \bm{\theta}, \bm{\eta} \right) p(\bm{v}'_n \mid \m{V} \setminus \{\bm{v}_n\}) p(\m{V} \setminus \{\bm{v}_n\})} {\tilde{f} \left( \x_n, {\lambda}_n; \bm{v}_n, s_n, \m{B}, \bm{\theta}, \bm{\eta} \right) p(\bm{v}_n \mid \m{V} \setminus \{\bm{v}_n\}) p(\m{V} \setminus \{\bm{v}_n\})} \frac {q(\bm{v}_n \mid \bm{v}_n')} {q(\bm{v}_n' \mid \bm{v}_n)} \\
&= \frac {\tilde{f} \left( \x_n, \lambda_{n}; \bm{v}_n', s_n', \m{B}', \bm{\theta}, \bm{\eta} \right)} {\tilde{f} \left( \x_n, \lambda_{n}; \bm{v}_n, s_n, \m{B}, \bm{\theta}, \bm{\eta} \right) } \,
\end{align*}
since $p(\m{B}) = q(\m{B})$, $p(\m{B}') = q(\m{B}')$ and
\[
p(\bm{v}_n \mid \m{V} \setminus \{\bm{v}_n'\}) = q(\bm{v}_n \mid \bm{v}_n') = \ncrp(\bm{v}_n; \m{V} \setminus \{\bm{v}_n'\}, \alpha) \, ,
\]
likewise for $p(\bm{v}'_n \mid \cdot)$ and $q(\bm{v}_n' \mid \bm{v}_n)$.

\section{Derivations for the Discriminant Variables}
\label{sec:rbhmc-sdv}
In order to derive the update steps for $\bm{\eta}$, one should focus on the associated term $(\lambda_{n} + C \rho_{n \ell z})^2$.
First, we derive
\begin{align*}
(\lambda_{n} + C \rho_{n \ell z})^2
&= (\lambda_{n} + C \varepsilon_{0} - C (\bm{\eta}_{v_{n \ell}} - \bm{\eta}_{z})^\top \x_n)^2 \\
&= (\lambda_{n} + C \varepsilon_{0})^2 - 2 C (\lambda_{n} + C\varepsilon_{0}) (\bm{\eta}_{v_{n \ell}} - \bm{\eta}_{z})^\top \x_n \\
&\quad + C^2 (\bm{\eta}_{v_{n \ell}}^\top \x_n - \bm{\eta}_{z}^\top \x_n)^2 \\
&= const. - 2 C (\lambda_{n} + C\varepsilon_{0}) \bm{\eta}_{v_{n \ell}}^\top \x_n + 2 C (\lambda_{n} + C\varepsilon_{0}) \bm{\eta}_{z}^\top \x_n \\
&\quad + C^2 \bm{\eta}_{v_{n \ell}}^{\top} \x_n \x_n^{\top} \bm{\eta}_{v_{n \ell}} - 2 C^2 \bm{\eta}_{v_{n \ell}}^{\top} \x_n \x_n^{\top} \bm{\eta}_{z} + C^2 \bm{\eta}_{z}^{\top} \x_n \x_n^{\top} \bm{\eta}_{z} \, .
\end{align*}

We point out that $\{n \mid v_{n \ell} = z\} \cap \{n \mid s_{n 2} = z\} = \emptyset$ according to our constraint set.
Hence, we need to sum up
\begin{align*}
&\sum_{n: (v_{n \ell} = z ~||~ s_{n 2} = z)} \lbr - \frac {(\lambda_{n} + C \rho_{n \ell s_{n 2}})^2} {2 \lambda_{n}} \rbr \\
&= \lcbr \sum_{n: v_{n \ell}=z} \frac 1 {\lambda_{n}} \lp \lp C {\lambda_{n}} + {C^2} \varepsilon_{0} \rp \x_n^{\top} + {C^2} \bm{\eta}_{s_{n 2}} \x_n \x_n^{\top} \rp \right. \\
&\quad \left. + \sum_{n: s_{n 2}=z} \frac 1 {\lambda_{n}} \lp - \lp C \lambda_{n} + C^2 \varepsilon_{0}\rp \x_n^{\top} + {C^2} \bm{\eta}_{v_{n \ell}} \x_n \x_n^{\top} \rp \rcbr \bm{\eta}_z \\
&\quad - \frac 1 2 \bm{\eta}_z^{\top} \lcbr C^2 \sum_{n} \mathbbm{1}(v_{n \ell} = z~||~s_{n 2} = z) \frac {\x_n \x_n^{\top}} {\lambda_{n}} \rcbr \bm{\eta}_z + const.
\end{align*}

Meanwhile, the canonical parameterization of a multivariate Gaussian distribution can be written as
\begin{align}
\N_{\rm canonical}(\bm{\eta}_z \mid \bm{\nu}_z, \m{\Lambda}_z) = \exp \lp const. + \bm{\nu}_z^{\top} \bm{\eta}_z - \frac 1 2 \bm{\eta}_z^{\top} \m{\Lambda}_z \bm{\eta}_z \rp \, .
\end{align}

Recall that we have assumed a prior $\N(\bm{0}, \nu_0^2 \m{I})$ for any $\bm{\eta}$.
Therefore, according to the canonical parameterization of the multivariate Normal distribution, we can derive
\begin{align*}
\bm{\nu}_z
&= \frac 1 {\lambda_{n}} \lcbr \sum_{n: v_{n \ell}=z} \left(\tilde{C}_n + C^2 \bm{\eta}_{s_{n 2}}^{\top} \x_n \right) - \sum_{n: s_{n 2}=z} \left( \tilde{C}_n - C^2 \bm{\eta}_{v_{n \ell}}^{\top} \x_n \right) \rcbr \x_n^{\top} \\
\m{\Lambda}_z
&= {C^2} \sum_{n} \mathbbm{1}(v_{n \ell} = z~||~s_{n 2} = z) \frac {\x_n \x_n^{\top}} {\lambda_{n}} + \nu_0^2 \m{I} \\
\tilde{C}_n &\coloneqq {C\lambda_{n}} + C^2 \varepsilon_{0} \, .
\end{align*}

\section{F-measure for Clustering}
\label{sec:f-measure}
Let $\{\bm{d}_1, \dots, \bm{d}_I\}$ be the set of $I$ ground truth classes from the data where $\bm{d}_i$ stores all data labeled with class $i$.
Apart from that, $\{\bm{\omega}_1, \dots, \bm{\omega}_J\}$ denotes the $J$ predicted clusters generated by some algorithm where $\bm{\omega}_j$ stores all data that are assigned to cluster $j$.
Let $N_{i j}$ be the number of datum overlapping among the ground truth class $i$ and the predicted cluster $j$ where $N_{i j} = \lvert \bm{d}_i \cap \bm{\omega}_j \rvert$.
Also, we denote $N_{i \cdot} = \sum_j N_{i j}$ and $N_{\cdot j} = \sum_i N_{i j}$.
This score is composed of precision (prec) and recall where
\begin{align*}
  \mathrm{prec}(i, j) = \frac {N_{i j}} {N_{\cdot j}} \qquad \mbox{and} \qquad \mathrm{recall}(i, j) = \frac{N_{i j}}{N_{i \cdot}} \, .
\end{align*}
Finally the measure is defined by
\begin{align*}
\mbox{F-measure} = \sum_i \frac{N_{i \cdot}}{n} \max_{j} \frac {2 \lbr \mathrm{recall}(i, j) \times \mathrm{prec}(i, j) \rbr}{\mathrm{recall}(i, j) + \mathrm{prec}(i, j)} \, .
\end{align*}

\section{Manual Labeling for \texttt{Animals}}
\label{sec:animal_label}
The table is shown in~\Cref{tb:animal_labels}.

\begin{table}[!ht]
\centering
\caption{Manual labels for \texttt{Animals}} \label{tb:animal_labels}
\begin{tabular}{lllll}
\toprule
   Animal &         Class & & Animal &         Class\\
\midrule
 Elephant &  land mammals & \quad \quad &     Seal & water animals \\
    Rhino &  land mammals & \quad \quad &    Dolphin & water animals \\
    Horse &  land mammals & \quad \quad &    Robin &         birds \\
      Cow &  land mammals & \quad \quad &   Eagle &         birds \\
    Camel &  land mammals & \quad \quad &    Chicken &         birds \\
  Giraffe &  land mammals & \quad \quad &    Salmon &          fish \\
  Gorilla &  land mammals & \quad \quad &    Trout &          fish \\
    Chimp &  land mammals & \quad \quad &   Bee &       insects \\
    Mouse &         mouse & \quad \quad &    Iguana &    amphibians \\
 Squirrel &         mouse & \quad \quad &    Alligator &    amphibians \\
    Tiger &     predators & \quad \quad &    Butterfly &       insects \\
      Cat &     predators & \quad \quad &    Ant &       insects \\
      Dog &     predators & \quad \quad &   Finch &         birds \\
     Wolf &     predators & \quad \quad &    Penguin & water animals \\
     Lion &     predators & \quad \quad &    Cockroach &       insects \\
     Whale & water animals & \quad \quad &  Ostrich &         birds \\
     Deer &  land mammals \quad \quad & & & \\
\bottomrule
\end{tabular}
\end{table}

\end{appendices}


\end{document}